\documentclass[letterpaper]{article} 
\usepackage{aaai2026}  
\usepackage{times}  
\usepackage{helvet}  
\usepackage{courier}  
\usepackage[hyphens]{url}  
\usepackage{graphicx} 
\usepackage{natbib}  
\usepackage{caption}
\usepackage{subcaption}
\usepackage{pgfplots} 
\usepackage[most]{tcolorbox}
\tcbuselibrary{skins, breakable, listings, theorems}
\usepackage{xcolor}
\usepackage{enumitem}

\usepackage{algorithm}
\usepackage{algorithmic}
\usepackage{subcaption}
\usepackage{amsmath}
\usepackage{amssymb}
\usepackage{amsfonts}
\usepackage{booktabs}
\usepackage{comment}
\newcommand{\myfirstpara}[1]{\par \noindent \textbf{#1}~}
\newcommand{\mypara}[1]{\par \noindent \textbf{#1:}~}

\def\vqa{\texttt{VQA}}

\usepackage{hyperref}
\usepackage{newfloat}
\usepackage{listings}
\DeclareCaptionStyle{ruled}{labelfont=normalfont,labelsep=colon,strut=off} 
\floatstyle{ruled}
\newfloat{listing}{tb}{lst}{}
\floatname{listing}{Listing}
\title{Refine and Align: \\ Confidence Calibration through Multi-Agent Interaction in VQA}
\author{
    Ayush Pandey\textsuperscript{\rm 1},
    Jai Bardhan\textsuperscript{\rm 1, \rm 2}\thanks{Work done while at TCS Research},
    Ishita Jain\textsuperscript{\rm 1},
    Ramya S Hebbalaguppe\textsuperscript{\rm 1},
    Rohan Raju Dhanakshirur\textsuperscript{\rm 1},
    Lovekesh Vig\textsuperscript{\rm 1}
}
\affiliations {
    \textsuperscript{\rm 1}TCS Research\\
    \textsuperscript{\rm 2}Czech Institute of Informatics, Robotics and Cybernetics, Czech Technical University in Prague \\
    ayush.p4@tcs.com,
    jai.bardhan@cvut.cz,
    ishita2403@gmail.com,
    ramya.hebbalaguppe@tcs.com,
    rohanrd@sit.iitd.ac.in,
    lovekeshvigin@gmail.com 
}

\def\cal{\textit{AlignCal}~}
\usepackage{bibentry}

\begin{document}

\maketitle
\begin{figure*}[t]           
  \centering
  \includegraphics[width=0.9\textwidth]{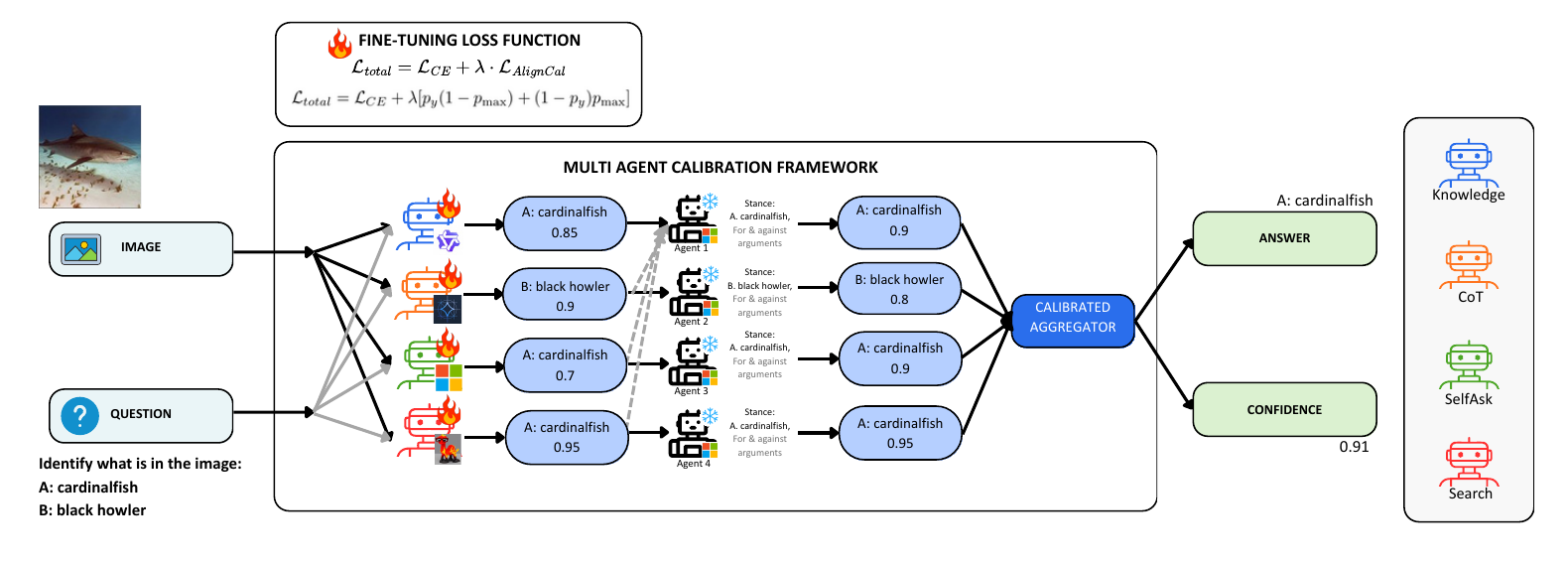}
  \caption{ \textbf{\textbf{AlignVQA} Multi-Agent Calibration Model.} Given an input image and question, the model first queries a set of specialized agents—\textit{Chain-of-Thought} \cite{wei2022chain}, \textit{Search-Augmented}, \textit{SelfAsk} \cite{press2022measuring}, and \textit{GENREAD Knowledge-based} \cite{yu2022generate} models—each fine-tuned for calibration using our custom proposed loss \cal. These agents independently produce answer classes (e.g., A: cardinalfish, B: black howler). In the second stage, a group of general agents is instantiated, with each agent probabilistically initialized to a specific answer class based on the distribution of predictions from the specialized agents. These general agents then receive argument-based feedback—comprising \textit{for} and \textit{against} justifications—from all general agents (denoted by dotted grey lines), enabling them to revise both their stance and confidence. The final calibrated prediction is the majority-vote class with associated confidence.
}
  \label{fig:wide_image}
\end{figure*}

\begin{abstract}

In the context of Visual Question Answering (VQA) and Agentic AI, calibration refers to how closely an AI system's confidence in its answers reflects their actual correctness. This aspect becomes especially important when such systems operate autonomously and must make decisions under visual uncertainty. While modern VQA systems, powered by advanced vision-language models (VLMs), are increasingly used in high-stakes domains like medical diagnostics and autonomous navigation due to their improved accuracy, the reliability of their confidence estimates remains under-examined. Particularly, these systems often produce overconfident responses. To address this, we introduce \textit{AlignVQA}, a debate-based multi-agent framework, in which diverse specialized VLM -- each following distinct prompting strategies -- generate candidate answers and then engage in two-stage interaction: generalist agents critique, refine and aggregate these proposals. This debate process yields confidence estimates that more accurately reflect the model’s true predictive performance. We find that more calibrated specialized agents produce better aligned confidences. Furthermore, we introduce a novel differentiable calibration-aware loss function called \cal designed to fine-tune the specialized agents by minimizing an upper bound on the calibration error. This objective explicitly improves the fidelity of each agent’s confidence estimates. Empirical results across multiple benchmark VQA datasets substantiate the efficacy of our approach, demonstrating substantial reductions in calibration discrepancies. Furthermore, we propose a novel differentiable calibration-aware loss to fine-tune the specialized agents and improve the quality of their individual confidence estimates based on minimising upper bound calibration error. Empirical evaluations across multiple benchmark VQA datasets demonstrate the effectiveness of our approach in significantly reducing calibration discrepancies.
\begin{enumerate}
    \item[] \textbf{Code}: \url{https://github.com/ayushp88/AgenticCalibration}
    \item[] \textbf{Website}: \url{https://refine-align.github.io/}
\end{enumerate}

\end{abstract}
\section{Introduction}

\myfirstpara{Visual Question Answering (\vqa)} is a foundational task in multimodal artificial intelligence that requires models to jointly process visual content and natural language to generate accurate answers to open-ended questions about images. First introduced to connect vision and language for goal-oriented reasoning \cite{antol2015vqa}, VQA has evolved into a benchmark for evaluating systems’ abilities in compositional reasoning, visual grounding, and language understanding \cite{agrawal2016vqavisualquestionanswering}. The task tests a model’s capability to extract relevant visual cues, understand complex queries, and synthesize answers—all in a single pipeline.

\mypara{Agentic architectures for \vqa} 
Recent advancements in VQA have embraced agentic architectures, where multiple interacting agents collaboratively solve complex visual reasoning tasks. For instance, Jiang et al. \cite{agentic_1} introduced a zero-shot multi-agent system with specialized experts coordinated adaptively. Hu et al. \cite{agentic_2} proposed a team of LLM-based agents with tool access, whose outputs are aggregated via voting. Wang et al. \cite{agentic_3} designed explainable agents with dedicated roles (Responder, Seeker, Integrator) that operate in a top-down reasoning loop. Similarly, the ARE model \cite{agentic_4} focuses on dynamic action-based reasoning through interaction-aware agents grounded in commonsense. However, these methods often rely on static interaction protocols or fixed agent roles, which limits their adaptability to new types of questions. Additionally, most do not explicitly optimize for confidence calibration, which is critical in high-stakes VQA applications.

\mypara{Need for Calibration in \vqa} Due to its practical relevance, VQA is increasingly being deployed in high-stake real-world domains such as medical diagnosis \cite{lin2023medical, zhou2023medicalvisualquestionanswering, canepa2023visualquestionansweringmedical}, autonomous navigation \cite{qian2024nuscenesqamultimodalvisualquestion, sima2025drivelmdrivinggraphvisual,marcu2024lingoqavisualquestionanswering,atakishiyev2023explainingautonomousdrivingactions}, and assistive technologies for the visually impaired \cite{gurari2018vizwizgrandchallengeanswering, chanana2017assistive}. In these settings, it is not only essential for VQA systems to be accurate, but also to be calibrated. A model is said to be calibrated if it's confidence matches the probablity of occurance \cite{guo2017calibration}. A calibrated model knows when to trust their predictions. Uncalibrated \vqa models pose serious safety threats. For example, an incorrect but overconfident answer in a medical application can mislead clinicians, while a misjudged scene description in autonomous driving could lead to hazardous decisions. In accessibility tools, a high confident false or misleading answers can reduce user trust posing safety risks.
Despite significant improvements in answer accuracy, many leading VQA models are often miscalibrated. i.e. they express high confidence even when their answers are wrong \cite{groot2024overconfidencekeyverbalizeduncertainty,zhang2024exploring}. This overconfidence undermines the reliability and interpretability of model predictions, especially when the stakes are high. 

\mypara{Measuring Calibration}
To evaluate calibration, standard tools such as reliability diagrams and scalar metrics like Expected Calibration Error (ECE) are widely used \cite{guo2017calibration, mce}. In this work, we focus on the multiple-choice (MCQ) VQA setting, where the model selects one answer from a predefined set. Calibration in this setting is measured using the standard ECE metric:
\begin{equation}\label{eq:ece_mcq}
ECE_{MCQ} := \sum_{m=1}^M \frac{\lvert B_m\rvert}{n}\bigl\lvert\mathrm{acc}(B_m) - \mathrm{conf}(B_m)\bigr\rvert
\end{equation}
Here, $B_m$ denotes the set of predictions with confidence values falling into the $m$-th bin, $\mathrm{acc}(B_m)$ and $\mathrm{conf}(B_m)$ are the average accuracy and confidence in the bin, and $n$ is the total number of samples. A well-calibrated model should produce predictions where the reported confidence matches the actual probability of correctness.

\mypara{Calibration in SOTA VQA architectures}
Several recent works have attempted to improve VQA calibration. Whitehead et al. \cite{whitehead2022reliable} proposed a selective answering strategy where the model abstains when it is unsure. Mozaffari et al.’s GLEN framework \cite{mozaffari2025glen} introduced a combination of model simplification and focal loss to enhance calibration. IVON by Wieczorek et al. \cite{wieczorek2025variational} leveraged Bayesian variational fine-tuning to capture model uncertainty through posterior weight distributions. While these methods are effective, they come with multiple limitations. Some of them require retraining or fine-tuning the model, while others operate in a single-pass setting without iterative refinement. The potential of multi-agent collaboration for post-hoc calibration remains largely unexplored in the VQA context.

\par\noindent\textbf{Human-Inspired Calibration via Multi-Agent Debate}
Humans rarely make decisions in isolation—opinions evolve through discussion, critique, and consensus. This notion of collective wisdom has been explored in recent AI literature through multi-agent debate frameworks that align confidence with justification through interaction \cite{yang2024confidence, oriol2025multiagentdebatestrategiesenhance, lin-etal-2022-truthfulqa}. Inspired by this, we introduce AlignVQA, a novel calibration method for MCQ \vqa. Our approach leverages a structured multi-agent debate, where specialized agents generate initial answers, and generalist agents critique, revise, and update both answers and confidences through a collaborative process. Towards this, we make the following contributions:

\begin{enumerate}
\item \textbf{Multi-Agent VQA Debate Framework.} We propose a structured post-hoc debate setup in which specialized agents generate initial predictions, and generalist agents iteratively critique and revise these answers and their confidences. The final answer is chosen using a confidence-based aggregation mechanism that maximizes the average confidence across agents. This leads to improved calibration and robustness. Upon applying the proposed agentic framework, the ECE on the VQARad dataset is substantially reduced from its initial value of 0.375 (as reported for Gemma 3 4B) to 0.146. Likewise, the Adaptive Calibration Error (ACE) decreases from 0.207 to 0.133, indicating improved calibration.

\item \textbf{Differentiable Calibration-Aware Loss:} 
We propose a novel differentiable loss function, termed \emph{AlignCal}, which is surrogate to minimize a provable upper bound on the ECE. Unlike conventional loss functions such as cross-entropy or focal loss, which primarily focus on prediction accuracy, \emph{AlignCal} jointly optimizes for both answer correctness and calibration, thereby enhancing the reliability of the specialized agents during training. 
\emph{AlignCal} leads to a reduction of ECE from 0.232 (focal loss finetuned Gemma 3 4B) to 0.058 on ScienceQA dataset. Incorporating the \emph{AlignCal} finetuned agents in debate framework leads to further reduction of ECE to 0.055 and ACE to 0.110.

\end{enumerate}
\begin{table*}[ht]
  \centering
  \small
  \setlength{\tabcolsep}{1mm}
    \begin{tabular}{l|
                    cccc ccc|
                    cccc ccc}
      \toprule
      {\textbf{Architecture}} 
        & \multicolumn{7}{c|}{\textbf{ScienceQA}} 
        & \multicolumn{7}{c}{\textbf{VQARad}} \\
      \cmidrule(lr){2-8} \cmidrule(l){9-15}
        & $\uparrow$\textbf{Acc.} 
        & $\uparrow$\textbf{F1} 
        & $\uparrow$\textbf{Prec.} 
        & $\uparrow$\textbf{Rec.} 
        & $\downarrow$\textbf{ACE} 
        & $\downarrow$\textbf{ECE} 
        & $\downarrow$\textbf{MCE} 
        & $\uparrow$\textbf{Acc.} 
        & $\uparrow$\textbf{F1} 
        & $\uparrow$\textbf{Prec.} 
        & $\uparrow$\textbf{Rec.} 
        & $\downarrow$\textbf{ACE} 
        & $\downarrow$\textbf{ECE} 
        & $\downarrow$\textbf{MCE} \\
      \midrule
      LLAVA One Vision 
        & 67.20\%  & 0.380 & 0.384 & \underline{0.405} & \underline{0.338} & \underline{0.335} & \textbf{0.366} 
        & 47.00\%  & 0.348 & 0.570 & 0.507 & 0.2312 & \underline{0.232} & \textbf{0.286} \\
      Gemma 3 4B  
        & \underline{71.00}\%  & \underline{0.443} & \textbf{0.582} & 0.372 & 0.398 & 0.398 & \underline{0.464} & \underline{59.40}\%  & \underline{0.591} & \underline{0.606} & \underline{0.601} & \underline{0.208} & 0.375 & 0.818 \\
      Qwen2.5-VL-3B-Inst. 
        & 69.70\%     & \textbf{0.482} & \underline{0.467} & \textbf{0.504} & \textbf{0.302} & \textbf{0.302} & 0.702 
        & \textbf{69.00\%} & \textbf{0.710} & \textbf{0.700} & \textbf{0.680} & 0.294 & 0.295 & \underline{0.297} \\
      Phi-4-multimodal-Inst. 
        & \textbf{76.00}\%  & 0.235 & 0.230 & 0.254 & 0.575 & 0.574 & 0.657 & 58.00\%  & 0.580 & 0.580 & 0.579 & \textbf{0.109} & \textbf{0.134}  & 0.425\\
      \bottomrule
    \end{tabular}%
    \caption{Comparison of state‐of‐the‐art model performance on the ScienceQA and VQARad datasets. Bold values indicate best performance for each metric within each dataset, underlined values indicate second-best performance.}
    \label{tab:sota_combined}
\end{table*}

\section{Related Works}
The common calibration techniques used in classification tasks include: (I.)  \textbf{Train-time Calibration methods} aim to improve confidence estimates during the training phase by modifying the loss function. These methods generally smooth confidence scores in a sample-agnostic manner--applying regularization uniformly across samples. For example, label smoothing \cite{szegedy2016rethinking} is a popular train-time calibration method which was originally proposed to improve classifier accuracy by computing the cross-entropy with a weighted sum of the one-hot vector with a uniform distribution. Other works include \cite{ ghosal2025better,Patra_2023_WACV,hebbalaguppe2022stitch,lin2017focal,hebbalaguppe2024calibration} (II.)\textbf{Post-hoc Calibration} Post-hoc calibration methods adjust a fully trained model’s confidence scores using a separate hold-out set. For example, \cite{guo2017calibration} introduced temperature scaling (TS), which smooths confidence by dividing the logits by a scalar T$>$1. Other notable post-hoc techniques include  \cite{bohdal2021meta,islam2021class,hebbalaguppe2025prompting}.

\mypara{Multi-Agent Calibration in LLM}
Collaborative Calibration \cite{yang2024confidence} was introduced as a multi-agent deliberation framework where agents share their predictions, confidence estimates, and the reasoning steps to engage in a simulated group dialogue. Agents iteratively refine responses, leading to improved calibration. This post-hoc ensemble method for LLMs inspires our vision-language adaptation. 
\textbf{Calibration in VQA}
 Utilizing a popular strategy of using consistency among samples to estimate confidence, Eisenschlos et al. \cite{eisenschlos2024selectively} introduced a method for improving the reliability of visual question answering (VQA) models. Their calibration approach involves generating multiple answers to a given question and computing the expected pairwise BLEU score, weighted by likelihood of each response under the model's sampling distribution. Then, this expected BLEU score is used to estimate the confidence.
The authors applied their approach both directly on images and on gold-standard image captions generated by human annotators. However, a notable improvement in calibration metrics was observed only when using captions, highlighting a key limitation of this method.\\
\textbf{Multi-Agent Calibration in Other Visual tasks}
An ensemble technique for Image Classification introduced by Schulze et al. \cite{schulze2025classifier} consists of attaching multiple independently trained classifier ``heads" to a shared, frozen backbone. The ensemble methods  are aggregated using three methods: simple averaging, majority voting, and the use of metamodelings--where the outputs of all heads serve as inputs to a separate model that produces the final prediction. Among these, metamodeling yielded significantly better performance, whereas averaging and majority voting showed minimal improvements. While this approach is computationally efficient and achieves better calibration metrics, its evaluation is limited to relatively small datasets such as CIFAR-100, raising questions about its scalability to larger and more complex benchmarks. Moreover, since metamodeling was the only aggregation strategy to yield meaningful gains, it suggests that the improvement may stem more from the aggregator itself than from the ensemble structure.A related approach in the domain of object detection is MoCaE: Mixture of Calibrated Experts \cite{oksuz2023mocae}, which aims to improve the accuracy of confidence estimates through expert ensembling. 
\\
\textbf{Multi-Agent approaches in VQA tasks}
Wang et al. \cite{wang2023towards} proposed a multi-agent architecture for VQA that draws inspiration from the human process of top-down reasoning—where individuals leverage prior knowledge and contextual cues to infer new information (e.g., predicting rain from observing cloudy skies). Their framework, referred to as Top-Down Reasoning, consists of three specialized agents: the responder, a vision-language model (VLM) that generates answers to visual questions; the seeker, which formulates relevant follow-up questions based on contextual understanding; and the integrator, which synthesizes insights from both agents to produce the final response. While the paper does not address calibration, its multi-agent framework can enhance answer reliability via improved confidence estimates. To our knowledge, no prior work leverages multi-agent methods for calibration in VQA.

\section{Proposed Methodology}

\subsection{Preliminaries}
Vision Language Models (VLMs) in VQA are empirically miscalibrated (see Tab. \ref{tab:sota_combined}), often exhibiting overconfident incorrect answers. To address this, we propose a two-stage calibration strategy built on agentic debate and refinement: (i) first stage of diverse expert answer generation and semantic clustering, and (ii) a second stage confidence refinement through deliberative generalist agents, followed by calibration aware aggregation. The overall design draws inspiration from recent multi-agent debate and refinement strategies applied to LLMs~\cite{yang2024confidence}, and adapts it a VQA setting with VLMs.  

The VQA task can be formulated as learning a function $f:\mathcal{I}\times\mathcal{Q}\rightarrow\mathcal{A}$ that, given an image $i\in\mathcal{I}$ and its associated question $q\in\mathcal{Q}$, predicts an answer $\hat{y}\in\mathcal{A}$ and confidence $p \in [0, 1]$.

\subsection{Agent Ensemble and Stance Generation}

Given an input image-question pair (for example, ``Identify the type of fish in the picture'' (c.f. Fig. \ref{fig:wide_image}), we first generate candidate answers from a diverse set of \emph{specalized expert agents}. Each agent is created with a different VLM backbone -- Qwen2.5‑VL‑3B‑Instruct \cite{bai2025qwen2}, Llava‑Onevision\cite{li2024llava}, Gemma 3 4B \cite{team2025gemma} and Phi‑4‑multimodal‑instruct \cite{abouelenin2025phi} -- and a distinct prompting strategy to encourage diverse reasoning\footnote{This agentic framework is model agnostic, any set of VLM backbones can be substituted in place of those used here.}.
Specifically, we employ: Chain-of-Thought prompting \cite{wei2022chain} for multi-hop reasoning, Self-Ask prompting \cite{press2022measuring} for recursive problem decomposition, Search-style prompting startegy to incorporate external retrieval cues and the GENREAD style prompting \cite{yu2022generate} for structured comprehension.

Each expert agent $i$ (independently) produces an output $v_i = (\hat{y}_i, p_i)$, where $\hat{y}_i$ is the answer string and $p_i$ is its sequence probability. We infer the confidence of the sequence through the geometric mean of probabilities of next-tokens generated. 
This serves as the initial confidence estimation for a candidate answer of a particular agent.
Because different agents may produce semantically equivalent but lexically different answers, we merge the semantically equivalent answers into $K$ \emph{semantically unique stances},
\[
    \{s_1, s_2, \dots, s_K\}, \qquad K \leq N
\]
where $N$ is the total number of individual 
responses across agents and prompting strategies, using GPT-3.5 judge following~\cite{tian2023just}.

For each stance \( s_k \), we define its index set as
\[
\mathcal{I}_k = \{ i \mid \hat{y}_i = s_k \}
\]
which denotes all positions \( i \) such that the \( i \)-th answer \( \hat{y}_i \) is equal to the stance \( s_k \). 
The \emph{frequency} of stance \( s_k \) is then,
\[
f_k = |\mathcal{I}_k|
\]
and the \emph{mean confidence} associated with that stance is computed as
\[
\bar{c}_k = \frac{1}{f_k} \sum_{i \in \mathcal{I}_k} c_i
\]
where \( c_i \) denotes the sequence probability of the \( i \)-th answer. 
Stage 1 thus yields a set of triplets, 
\[
\{(s_k, f_k, \bar{c}_k)\}_{k=1}^K,
\] 
capturing both the diversity of opinion and its strength in terms of support and confidence.

\paragraph{Illustrated Failure Case.} Fig.~\ref{fig:wide_image} in the first stage, three agents each give the answer \emph{cardinal fish} with confidence scores of $0.85$, $0.70$, and $0.95$, producing an average confidence of $0.83$ for that stance, while a fourth agent answers \emph{black howler} with confidence $0.90$. A majority confidence based system would adopt \emph{black howler} as the consensus after the first stage, despite it being incorrect and lacking group support. Stage 2, is designed to revisit and refine such consensus through deliberation and counter argumentation.


\subsection{Group Debate with Rationale and Feedback}

The second stage introduces a set of \textbf{generalist deliberation agents} (no specialized prompting) whose role is to critically examine, defend and revise the candidate stances produced in Stage 1, forming a structured debate ensemble. To maintain the prior group consensus while allowing contrarian exploration, each generalist agent $j$ is assigned a stance $s_j$ by sampling proportionally to the frequencies $f_k$, i.e., $\text{Pr}(s_j = s_k) \propto f_k$. This maintains a soft bias towards majority supported views while still allowing minority stances to be reconsidered.


Each agent then argues for its stance by exploring diverse reasoning and developing rationales for defending it. Each reasoning path is unique and develops an ensemble of rationales for a particular stance. Agents then provide ratings and feedback to each rationale in terms of logical consistency, factuality, clarity and conciseness. Specifically, Chain-of-Verification style prompting~\cite{dhuliawala-etal-2024-chain} is used to check the factuality by generating underlying premises or assumptions. These premises are then further checked with a search augmented agent to identify unfactual statements in the feedback. 

Each general agent then receives a pair of arguments one sampled from the set of supporting arguments and one sampled from one of the opposing sides to form a debate pair. This mirrors two-sided deliberation paradigms in multi-agent reasoning and debate systems. Based on these arguments, each agent produces a final answer that incorporates the provided opposing argument, supporting argument and its previously assigned stance. Thus, the final answer is given by $y_j' = f_j(s_j, \bar{c}_j, a_p, a_n)$, where $a_p, a_n$ are the supporting and opposing arguments with ratings and feedback, and $s_j, \bar{c}_j$ is the initial stance with its associated confidence assigned to agent $j$. We also record the sequence probability of each agent’s final response \( y_j' \), which serves as the refined confidence score $\mathrm{Conf}(y_j')$.



After collecting the set of refined outputs, we $\{(y_j', \mathrm{Conf}(y_j'))\}_{j=1}^M$ from $M$ generalist agents, we aggregate it in two steps: First for each stance $s$, we define the agent index set $\mathcal{I}'_s = \{j \mid y_j' = s \}$ and update $f_k' = |\mathcal{I}'_k|$, and compute the mean refined confidence as
\[
\widehat{c}_k = \frac{1}{|\mathcal{I}'_k|} \sum_{j \in \mathcal{I}'_k} \mathrm{Conf}(y_j').
\]
The final answer is selected by majority vote by choosing the stance with the most supporting agents:
\begin{equation}
  s^* \;=\;\arg\max_{k \,\in\,\{1,\dots,K\}} f_k'
  \quad
\end{equation}
The final confidence is the mean confidence of the agents supporting the chosen stance. This aggregation yields a better indication of prediction, by weighing different arguments through deliberation. To further improve calibration, we introduce the following loss.

\subsection{Calibration Aware Finetuning}
\label{sec:aligncal}
Our system can benefit from better calibrated VLMs. Therefore, we introduce a novel surrogate loss function that directly minimizes a tight upper bound on miscalibration during training, thereby avoiding the pitfalls of post-hoc fixes. 
Classical metrics like ECE average confidence–accuracy gaps over broad bins, so they often fail to estimate the reliability of a single test example or a specific subpopulation. 
Consider a tumor screening classifier that reports confidence 0.9 on all five patients: it is correct on three cases (two tumor positives and one healthy negative) and incorrect on two (one missed tumor and one false positive), so the empirical accuracy is \(3/5=0.6\). Because all predictions fall into the same confidence bin, the standard ECE is \(|0.9-0.6|=0.3\), which reflects only the coarse aggregate gap. In contrast, the Upper Bound Calibration Error (UBCE)~\cite{ubce} averages per-instance absolute gaps: correct cases contribute \(1-0.9=0.1\) each and incorrect ones contribute \(0.9\) each, yielding 
\(
\text{UBCE} = (3\times0.1 + 2\times0.9)/5 = 0.42.
\) The ECE therefore understates the expected misalignment because it conflates the high-confidence errors with high-confidence correct predictions via binning, whereas UBCE exposes the full error—including the overconfidence on the incorrect tumor diagnosis—by aggregating each individual confidence–correctness gap.

Therefore, we construct a differentiable loss by applying the plug-in principle to the sample estimate of UBCE (an upper bound on calibration metrics). Minimizing this surrogate loss directly drives down ECE and provides gains to MCE.


\begin{table*}[htp]
  \centering
  \small
    \setlength{\tabcolsep}{1mm}
    \begin{tabular}{l|ccccccc|ccccccc}
      \toprule
       & \multicolumn{7}{c|}{\textbf{VQARad Dataset}} & \multicolumn{7}{c}{\textbf{ScienceQA Dataset}} \\
      \cmidrule{2-8} \cmidrule{9-15}
      \textbf{Architecture} & \textbf{$\uparrow$Acc.} & \textbf{$\uparrow$F$_1$} & \textbf{$\uparrow$Prec.} & \textbf{$\uparrow$Rec.} & \textbf{$\downarrow$ACE} & \textbf{$\downarrow$ECE} & \textbf{$\downarrow$MCE} & \textbf{$\uparrow$Acc.} & \textbf{$\uparrow$F$_1$} & \textbf{$\uparrow$Prec.} & \textbf{$\uparrow$Rec.} & \textbf{$\downarrow$ACE} & \textbf{$\downarrow$ECE} & \textbf{$\downarrow$MCE} \\
      \midrule
      Agentic Framework
        & 65.70\% & 0.540 & \underline{0.554} & 0.544 & 0.144 & 0.146 & 0.820 & 72.80\% & 0.340 & 0.346 & 0.328 & 0.265 & 0.270 & 0.438 \\
        \addlinespace[2pt]
      \hline 
      \addlinespace[2pt]
      \multicolumn{1}{l|}{\textbf{Post-Hoc Calibration}} & \multicolumn{7}{l|}{} & \multicolumn{7}{l}{} \\
      Agentic + TS
        & 65.70\% & 0.540 & \underline{0.554} & 0.544 & 0.114 & 0.117 & 0.765 & 72.80\% & 0.340 & 0.346 & 0.328 & 0.255 & 0.268 & \underline{0.421} \\
      Agentic + DC
        & 65.70\% & 0.540 & \underline{0.554} & 0.554 & \underline{0.097} & \textbf{0.041} & \textbf{0.113} & 72.80\% & 0.340 & 0.346 & 0.328 & -- & -- & -- \\
      \addlinespace[2pt]
      \hline
      \addlinespace[2pt]
      \multicolumn{1}{l|}{\textbf{Train-Time Calib.}} & \multicolumn{7}{l|}{} &  \multicolumn{7}{l}{} \\
      Agentic + FL
        & \textbf{68.50\%} & \underline{0.571} & 0.542 & \underline{0.605} & 0.116 & \underline{0.073} & 0.393 & 74.40\% & 0.424 & 0.480 & 0.381 & \underline{0.142} & \underline{0.180} & 0.678 \\
      Agentic + LS
        & 67.70\% & \textbf{0.650} & \textbf{0.652} & \textbf{0.650} & 0.175 & 0.183 & 0.543 & \underline{75.20\%} & \underline{0.467} & \underline{0.532} & \textbf{0.424} & 0.186 & 0.186 & 0.916 \\
      \addlinespace[2pt]
      \hline
      \addlinespace[2pt]
      \multicolumn{1}{l|}{\textbf{Proposed Method}} &  \multicolumn{7}{l|}{} & \multicolumn{7}{l}{} \\
      Agentic+\cal+FL
        & \underline{68.20\%} & 0.548 & 0.517 & 0.583 & \textbf{0.095} & 0.098 & \underline{0.267} & \textbf{76.10\%} & \textbf{0.472} & \textbf{0.540} & \underline{0.418} & \textbf{0.110} & \textbf{0.055} & \textbf{0.331} \\
      \bottomrule
    \end{tabular}%
  \caption{Comprehensive comparison of calibration strategies across VQARad and ScienceQA datasets. Bold values indicate best performance for each metric within each dataset, underlined values indicate second-best performance. The proposed method (Agentic + \cal + FL) demonstrates superior calibration performance with competitive accuracy across both datasets. It is not possible to perform Dirchelet Calibration in the case of ScienceQA Dataset due to the unavailability of the probabilities of other options.}
  \label{tab:calibration-comparison-merged}
\end{table*}

\subsubsection{Our loss function \textit{AlignCal}:}

Formally, given softmax outputs $\mathbf{p} = (p_1, \dots, p_K)$ with logits $z_i$, true label $y$, top predicted confidence $p_{\max} = \max_i p_i$ and predicted ground truth class probability $p_y$, we define the soft-calibration loss:
\begin{equation}
    \mathcal{L}_{\text{\cal}}(p_y, p_{\max}) = p_y(1-p_{\max}) + (1 - p_y)p_{\max}
\end{equation}

The full training objective therfore becomes:
\[
    \mathcal{L}_{\text{tot}} = \mathcal{L}_{\text{FL}} + \lambda \mathcal{L}_{\text{\cal}},
\]
where $\lambda$ is a tuneable hyperparameter and $\mathcal{L}_{\text{FL}}$ is the focal loss~\cite{mukhoti2020calibrating} which is a strictly proper scoring rule that encourages accuracy and also is found to implicitly enhance calibration. Our calibration term is not an ad hoc tweak; it arises naturally as a plug-in surrogate to UBCE:
\begin{multline*}
    \text{UBCE} = \text{Pr}(t=0)\mathbb{E}\left[p_{\max} \middle| t = 0\right] \\+ \text{Pr}(t=1)\left(1 - \mathbb{E}\left[p_{\max} \middle| t= 1\right]\right),
\end{multline*}
where $t=\mathbb{I}\{\hat{y} = y\}$ is an indicator function for whether the prediction is correct and $\text{Pr}(\cdot)$ defines the probability. Equivalently, algebraic regrouping gives us, 
\[
    \text{UBCE} = \mathbb{E}\left[t(1 - p_{\max}) + (1 - t)p_{\max}\right].
\]
This form makes it clear that UBCE is the expected absolute gap between the correctness and confidence -- hence a conservative upper bound on the calibration metrics. Applying the law of total expectation (tower rule/law of iterated expectations) and conditioning on an input $x$, we have,
\begin{align*}
    \text{UBCE} &= \mathbb{E}_x\left[\mathbb{E}\left[t (1 - p_{\max}) + (1 - t)p_{\max} \middle| x  \right]\right] \\
    &= \mathbb{E}_x\left[\mathbb{E}\left[t \middle| x\right] (1 - p_{\max}) + \mathbb{E}\left[(1 - t) \middle| x \right]p_{\max} \right].
\end{align*}
Here the outer expectation is over inputs \(x\) drawn from the data distribution, and the inner expectation is over the randomness in the correctness indicator $t$ given $x$ (i.e., over \(y\sim P(y\mid x)\)). Since \(p_{\max}\) is deterministic conditioned on \(x\), it can be pulled outside the inner expectation.

Let $q(x) := \mathbb{E}\left[t \middle| x\right]$ be the true conditional probability of correctness given an input $x$. Then,
\begin{equation*}
    \text{UBCE} = \mathbb{E}_x \left[q(x) (1 - p_{\max}) + (1 - q(x)) p_{\max} \right].
\end{equation*}

In practice, $q(x)$ is unknown because it depends on a (possibly deterministic) decision rule, such as an indicator function $\mathbb{I}\{\hat{y} = y\}$ which is non-differentiable and does not allow backpropagation. To obtain a differentiable surrogate, we plug in the model's own soft belief about correctness, namely, instead of taking $\hat{y} = \text{arg} \max p_i$, imagine sampling a label $\hat{y} \sim p\left(\cdot \middle| x \right)$ from the model's softmax output. Then the indicator function $t = \mathbb{I}\{\hat{y} = y\}$ has $\mathbb{E}\left[t \middle| x\right] = p_y$, i.e., the probability that a randomly drawn label equals the true label is exactly $p_y$. This makes $p_y$ a natural, smoothed estimator of $q(x)$. This replacement, is an instance of the classical plug-in principle for conditional expectation estimation -- which is well studied in statistical learning theory~\cite{plugin_estimators}. Accordingly, we define our soft surrogate loss per example as,
\begin{equation}
    \mathcal{L}_{\text{\cal}} (p_y, p_{\max}) = p_y (1 - p_{\max}) + (1 - p_y) p_{\max} \label{eq:l_cal}
\end{equation}
so that,
\begin{align*}
    \mathbb{E}\left[\mathcal{L}_{\text{\cal}}\right] &= \mathbb{E}_x \left[p_y(1-p_{\max}) +(1-p_y)p_{\max}\right] \\
    &\approx \text{UBCE},
\end{align*}
with equality in expectation under the assumption that $p_y \approx \mathbb{E}\left[t \middle| x\right]$. This construction gives a smooth, differentiable surrogate to UBCE that admits gradient-based optimization while preserving the structure of the original bound.

From a probabilistic calibration perspective, if the model is well calibrated in the sense that its softmax outputs reflect the true posterior -- i.e, $p_y \approx \text{Pr}(y|x)$ -- then $p_y$ also approximates $\text{Pr}(\hat{y} = y | x)$, making the plug-in substitution for $q(x)$ increasingly accurate. Finally, there is a self-correcting feedback: early in training $p_y$ might poorly estimate $q(x)$, but the calibration loss penalizes discrepancies between $p_{\max}$ and this current belief. Improving those discrepancies tends to make $p_y$ ``more honest'' about the correctness, which in-turn makes the surrogate tighter. This is analogous to consistency of plug-in classifier -- if the estimation of the conditional quantity improves, the overall decision or loss converges to the ideal~\cite{plugin_estimators}.

\subsubsection{Gradient Analysis of the proposed loss:}
To understand the learning dynamics induced by the combined objective, we analyze the gradients. We see that 
 \[
     \frac{\partial\mathcal{L}_{\text{\cal}}}{\partial z_i} = (1 - 2p_y)p_{\hat{y}}(\delta_{i,\hat{y}} - p_i) + (1 - 2p_{\hat{y}})p_y(\delta_{i,y} - p_i),
 \]
where $\hat{y} = \text{arg} \max_j p_j$ is the highest predicted class index, and $\delta_{i,j}$ is the Kronecker delta. This expression never collapses for the interior probabilities and thus provides a constant margin based push. When the model makes an overconfident incorrect guess, the loss acts to reduce the confidence on the erroneous top prediction and boosts the true class; when the model is underconfident, the calibration term works in conjunction with the cross-entropy term (focal loss) to increase $p_y$ above the threshold. We can further study the equilibrium case with respect to the loss to find that for a reasonable choice of $\lambda$, the combined loss provides a strict gradient for driving $p_y \rightarrow 1$, and for cases where $\hat{y} \neq y$, a gradient for driving $p_{\hat{y}} \rightarrow 0$.

\begin{figure*}[t]
  \centering
  \begin{subfigure}{0.19\textwidth}
    \includegraphics[width=\linewidth]{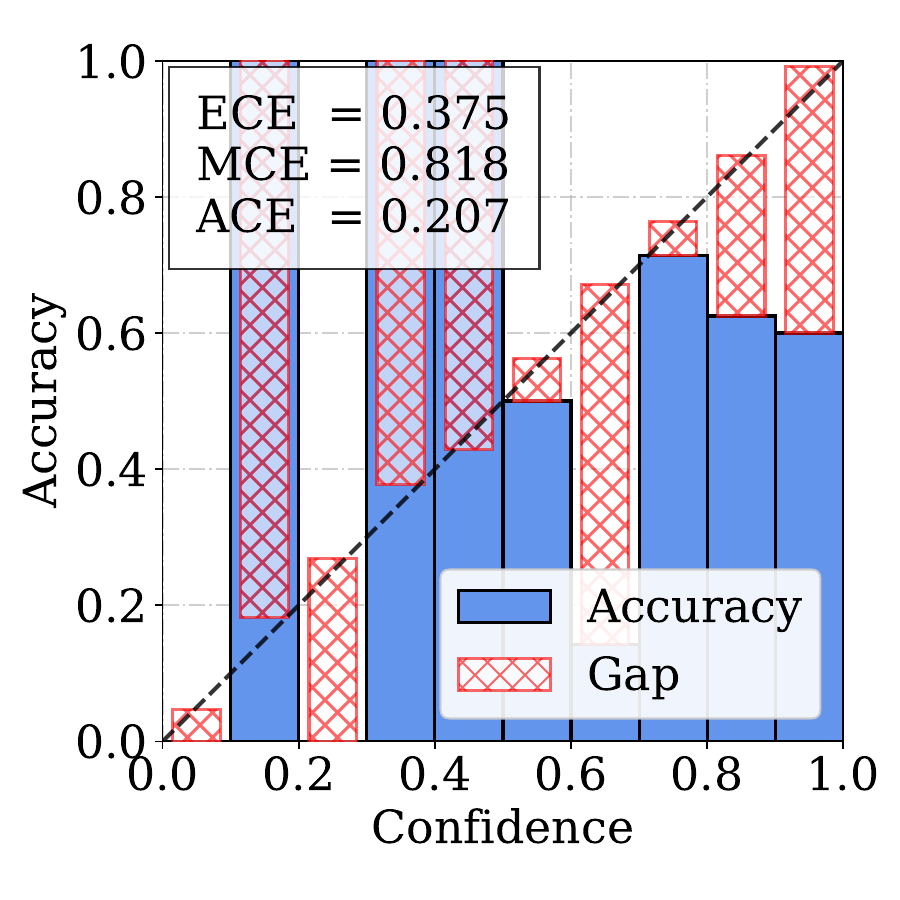}
  \end{subfigure}\hfill
  \begin{subfigure}{0.19\textwidth}
    \includegraphics[width=\linewidth]{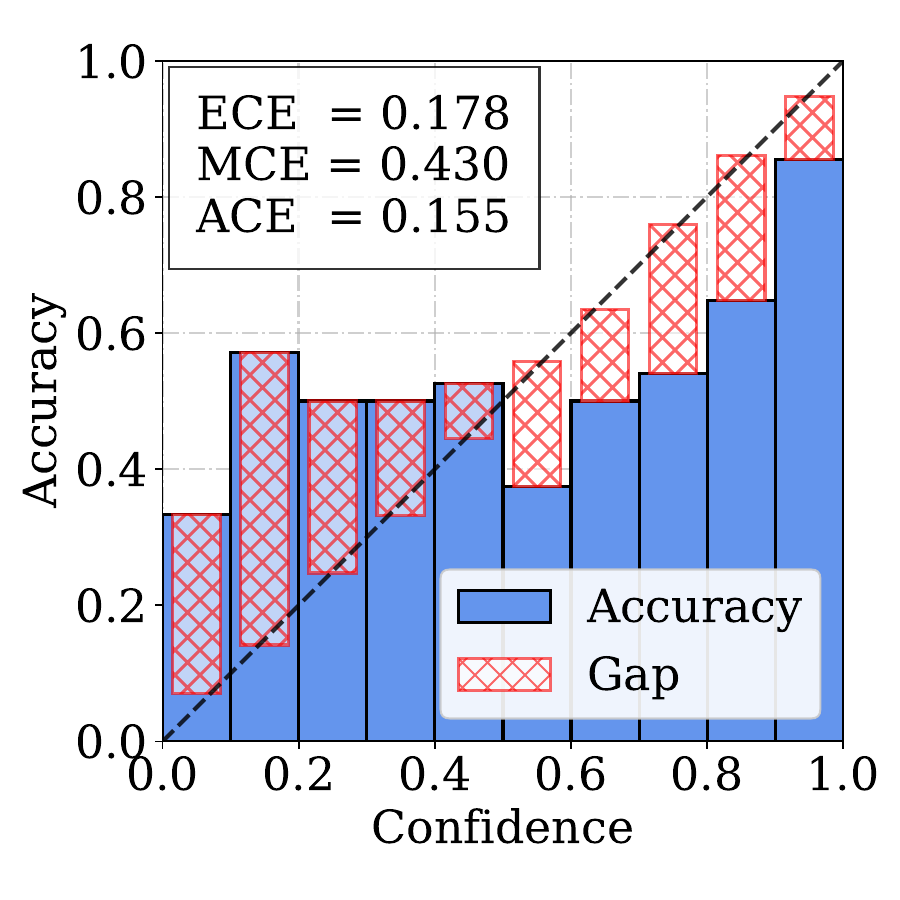}
  \end{subfigure}\hfill
  \begin{subfigure}{0.19\textwidth}
    \includegraphics[width=\linewidth]{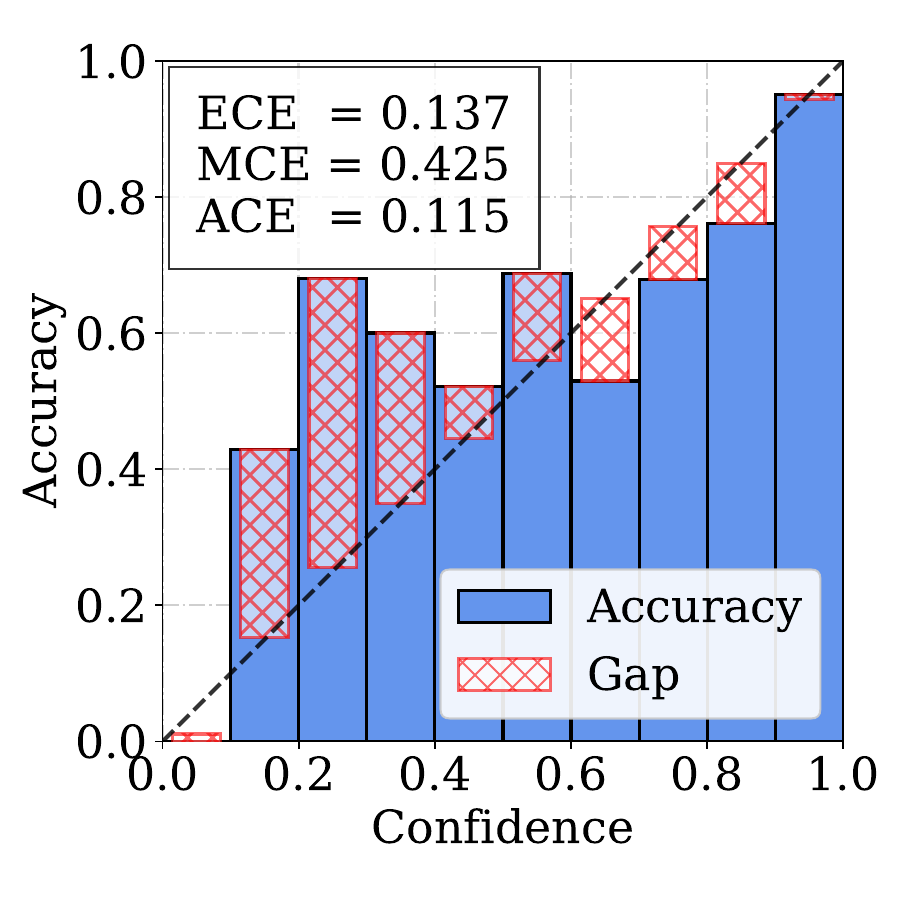}
  \end{subfigure}\hfill
  \begin{subfigure}{0.19\textwidth}
    \includegraphics[width=\linewidth]{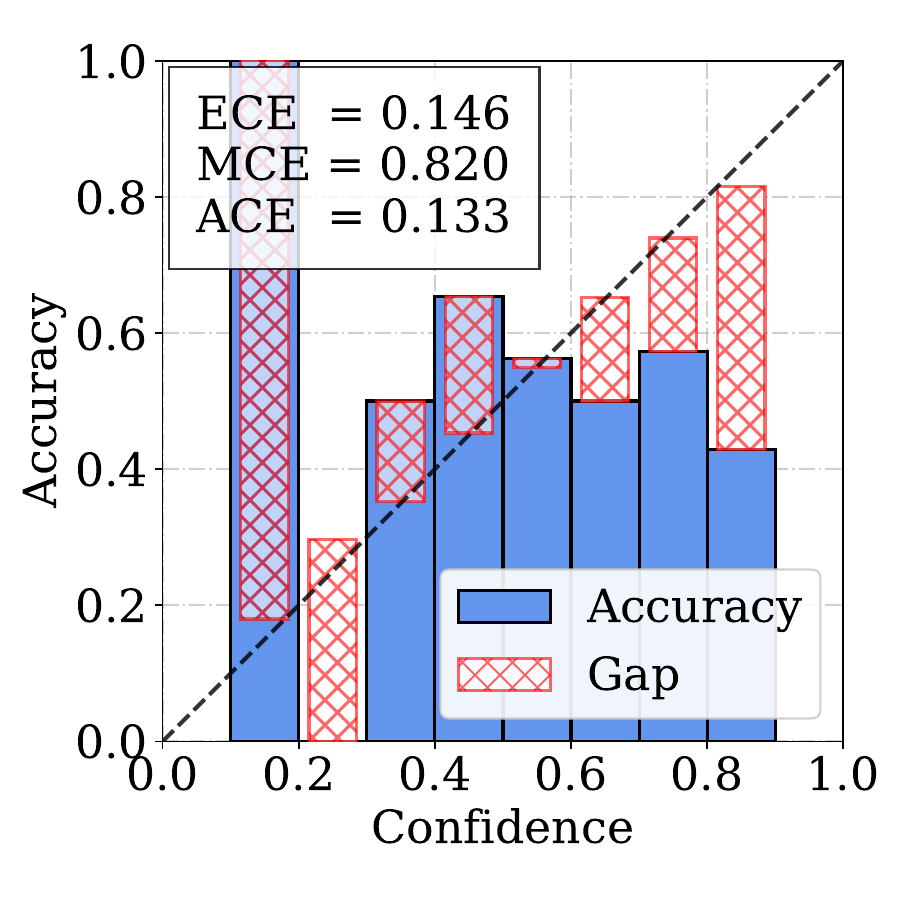}
  \end{subfigure}\hfill
  \begin{subfigure}{0.19\textwidth}
    \includegraphics[width=\linewidth]{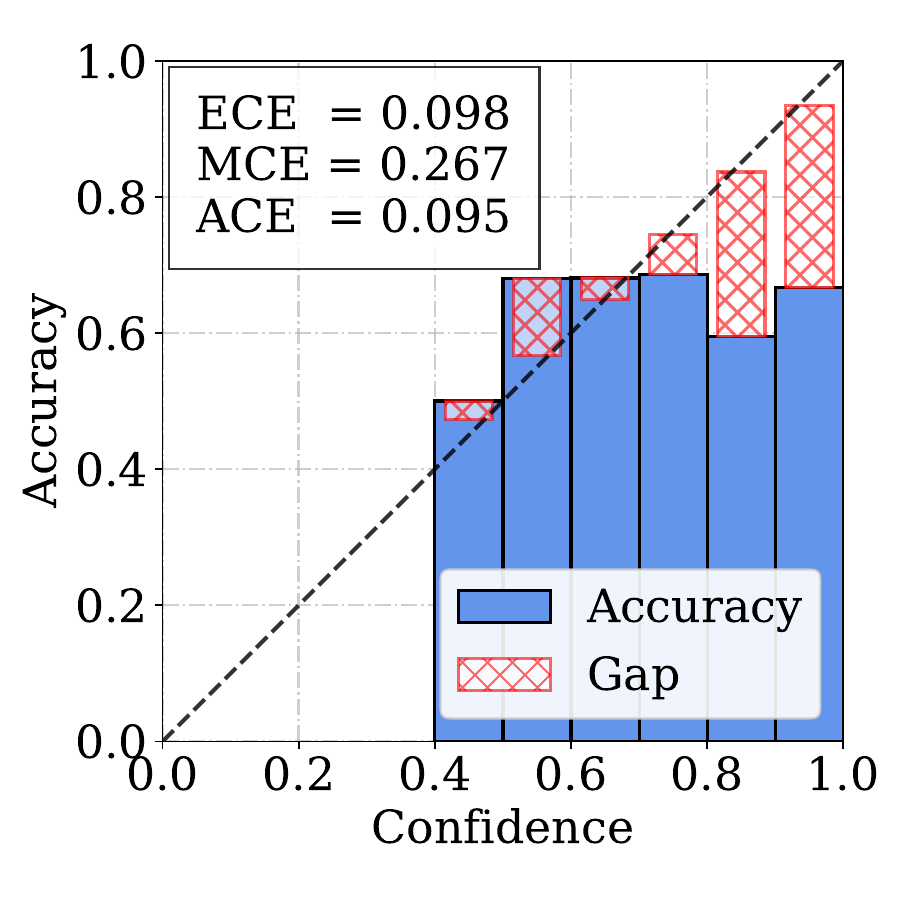}
  \end{subfigure}

  \setcounter{subfigure}{0}

  \begin{subfigure}{0.19\textwidth}
    \includegraphics[width=\linewidth]{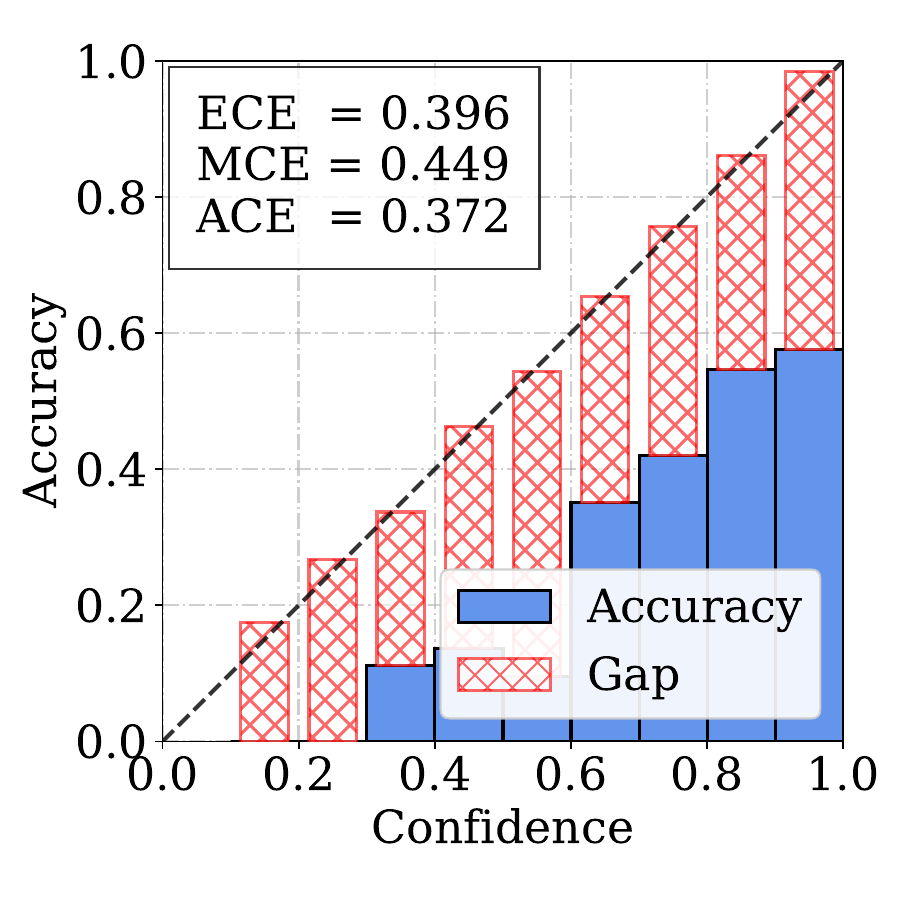}
    \caption{Gemma 3 4B}
    \label{fig:a-gemma}
  \end{subfigure}\hfill
  \begin{subfigure}{0.19\textwidth}
    \includegraphics[width=\linewidth]{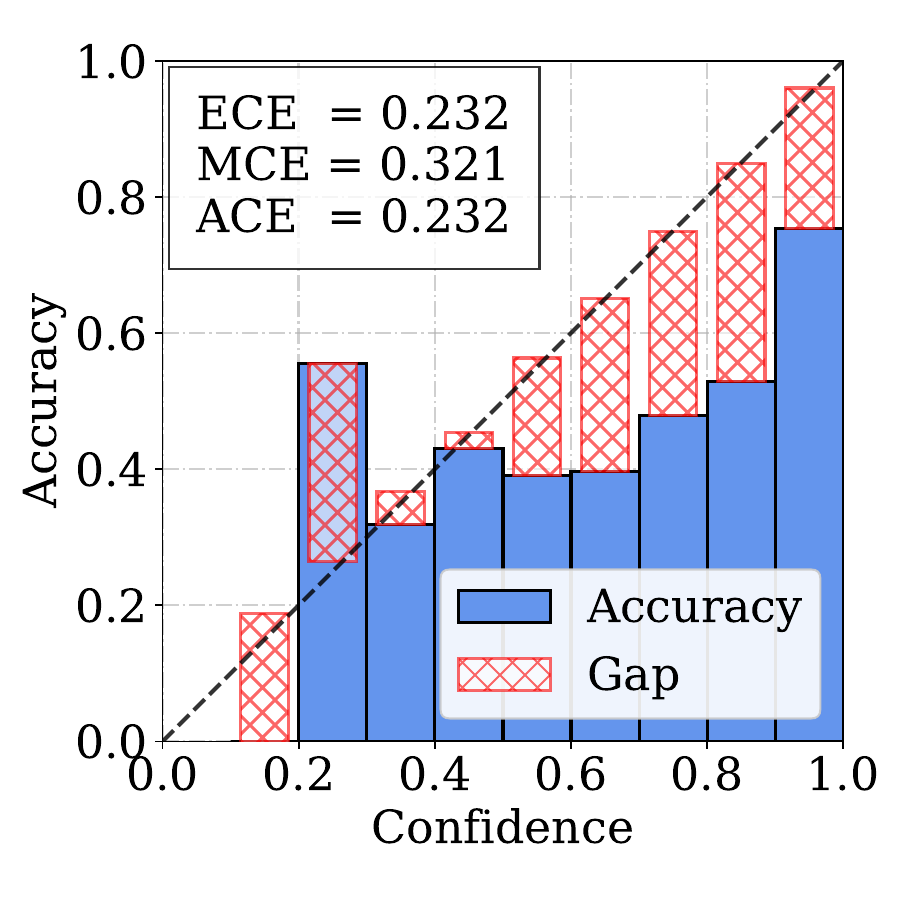}
    \caption{FL}
    \label{fig:b-fl}
  \end{subfigure}\hfill
  \begin{subfigure}{0.19\textwidth}
    \includegraphics[width=\linewidth]{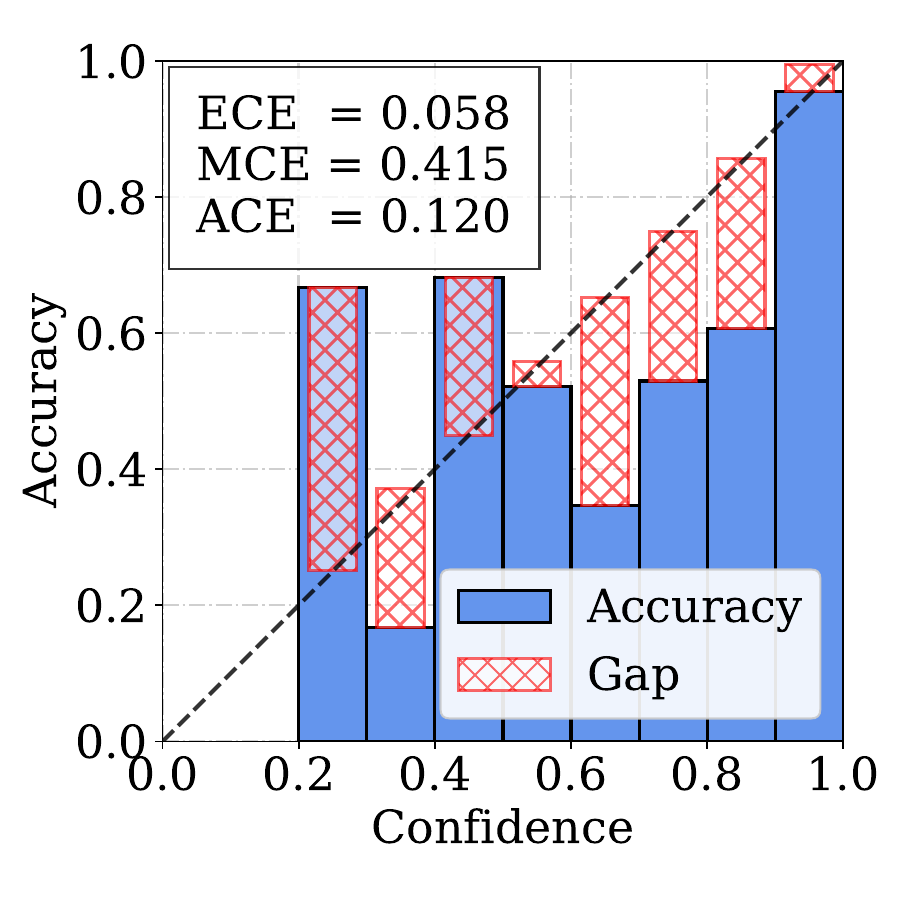}
    \caption{\cal + FL}
    \label{fig:c-flalign}
  \end{subfigure}\hfill
  \begin{subfigure}{0.19\textwidth}
    \includegraphics[width=\linewidth]{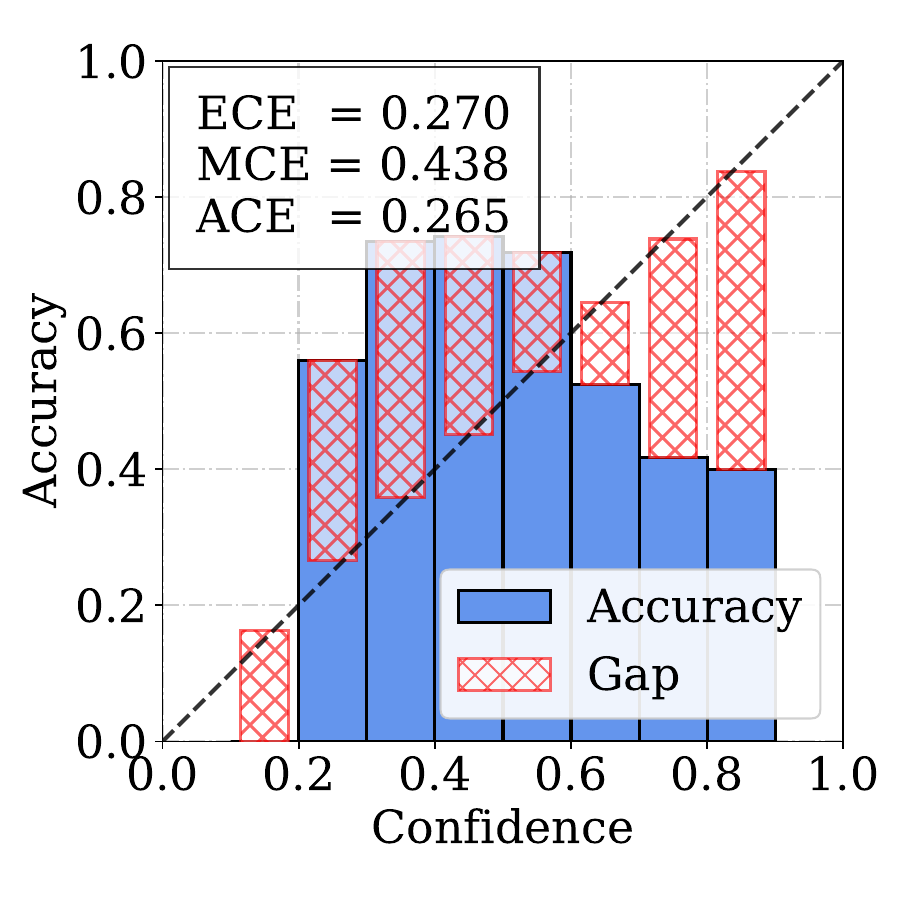}
    \caption{Agentic Framework}
    \label{fig:d-agentic}
  \end{subfigure}\hfill
  \begin{subfigure}{0.19\textwidth}
    \includegraphics[width=\linewidth]{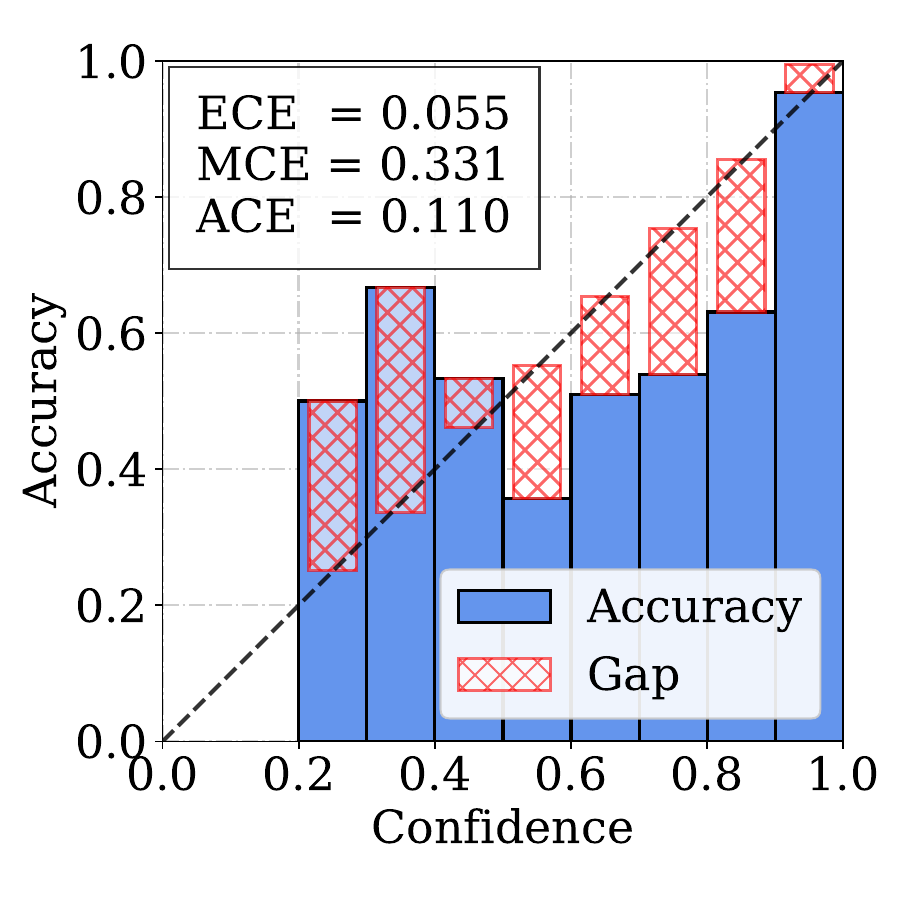}
    \caption{Agentic + \cal}
    \label{fig:e-agenticalign}
  \end{subfigure}

  \caption{Reliability plots of the datasets, VQARad (top) and ScienceQA (bottom). \ref{fig:a-gemma} shows the calibration from base Gemma model. \ref{fig:b-fl} shows plot on FL finetuned Gemma 3 4B model. \ref{fig:c-flalign} shows the plot on FL + \cal finetuned
Gemma 3 4B model. \ref{fig:d-agentic} shows the plot obtained from  Agentic Framework. \ref{fig:e-agenticalign} shows the plot obtained from
Agentic framework where agents are finetuned with \cal + FL.}
  \label{fig:combined-calibration}
\end{figure*}

Thus minimizing $\mathcal{L}_{\text{\cal}}$ directly tightens a provable upper bound on calibration error. 
Unlike MMCE \cite{kumar2018trainable} and other kernelized calibration penalties that require careful choice of kernels and suffer from increased per-update cost, our formulation is closed-form, hyperparameter-light (just $\lambda$). In contrast to label smoothing, which bluntly softens all targets and can undermine the sharpness of well-calibrated predictions, our loss targets the disparity between $p_{\max}$ and $p_y$, dynamically adjusting based on the model’s confidence geometry. Focal loss improves calibration only indirectly by reweighting hard examples for better accuracy, while our loss has an explicit probabilistic interpretation as a surrogate to UBCE, ensuring that reducing the training objective systematically reduces ECE by design. Worst-case deviation (MCE) is not guaranteed without extra uniformity constraints since we are minimizing an expectation. Though, we observe practical improvements there as well.
Our loss improves over the standard focal loss, as shown in Fig.~\ref{fig:combined-calibration}.

\section{Dataset and Evaluation}
\textbf{Datasets and Models:} We took 2 publicly available datasets, ScienceQA \cite{lu2022learn} and medical dataset VQARad \cite{lau2018dataset}. ScienceQA consists of 21,208 multimodal multiple choice questions with diversE science topics and annotations of their answers with corresponding lectures and explanations. VQA-RAD is manually constructed dataset in radiology where answers about images are naturally created and validated by clinicians. VQA-RAD dataset contains 3,515 total visual questions. We only consider Yes/No type of questions from this dataset. 
The agentic framework consists of four VLM backbones Qwen2.5‑VL‑3B‑Inst. \cite{bai2025qwen2}, Llava‑Onevision \cite{li2024llava}, Gemma 3 4B \cite{team2025gemma}, and Phi‑4‑multimodal‑Inst. \cite{abouelenin2025phi}. For generalist agent Phi‑4‑multimodal‑instruct, is taken as backbone.

\noindent \textbf{Evaluation}: We evaluate calibration with ECE \cite{guo2017calibration}, ACE, and MCE. To visualize miscalibration, we include reliability diagrams.  For task performance, we additionally report Accuracy, F1‑score, Precision, and Recall.
Maximum Calibration Error (MCE) is the largest absolute difference between predicted confidence and empirical accuracy across all confidence bins.
\begin{equation}
\mathrm{MCE}
= \max_{b=1,\dots,B}
  \bigl\lvert \mathrm{acc}(B_{b}) - \mathrm{conf}(B_{b}) \bigr\rvert
\label{eq:mce}
\end{equation}
Adaptive Calibration Error (ACE) splits the sorted predictions into bins each containing an equal number of examples, then computes the mean absolute gap between empirical accuracy and average confidence across those bins.
\begin{equation}
\mathrm{ACE}
= \frac{1}{B} \sum_{b=1}^{B}
  \bigl\lvert \mathrm{acc}(B_{b}) - \mathrm{conf}(B_{b}) \bigr\rvert
\label{eq:ace}
\end{equation}
\textbf{Comparison:} We compare our multi-agent debate framework (viz. \textit{Agentic Framework}) to the base VLM models, and compare our loss \cal against other standard calibration techniques Focal Loss (FL) \cite{mukhoti2020calibrating}, Label Smoothing (LS) \cite{szegedy2016rethinking}, and post‑hoc methods, including temperature scaling (TS) \cite{guo2017calibration} and Dirichlet calibration (DC) \cite{kull2019beyond}. 
\section{Experiment and Results}

\textbf{Agentic Framework} We report here the Agentic framework results. From Fig. \ref{fig:combined-calibration}, we can show that agentic VLM debate leads to better calibration and more reliable responses. On ScienceQA dataset ECE decreases from 0.396 to 0.270, while MCE and ACE also show consistent reductions from 0.449 to 0.438 and 0.372 to 0.265, respectively. In the VQARad dataset, ECE decreases from 0.375 to 0.143, ACE also shows consistent reduction 0.207 to 0.144. 

\noindent \textbf{\cal Results.} Fig. ~\ref{fig:combined-calibration} reports the calibration improvements achieved by \cal on two VQA benchmarks. On the VQARad dataset, ECE decreases from $0.178$ to $0.137$, and ACE decreases from $0.155$ to $0.115$. Similarly, on the ScienceQA dataset, ECE is reduced from $0.232$ to $0.058$, while ACE falls from $0.232$ to $0.120$. We also compared our results with other training calibration methods FL\cite{lin2017focal} and label smoothing \cite{szegedy2016rethinking}. For LS, we use $\alpha$ = 0.1 and for FL, we use $\gamma$ = 2.

\noindent \textbf{Comparison with Train Time and Post Hoc Calibration techniques}. We apply temperature scaling and dirchelet calibration on the results obtained from the Baseline agentic setup (Table \ref{tab:calibration-comparison-merged}). ECE improves from 0.1430 to 0.1165, while MCE improves from 0.82 to 0.7634 on applying temperature scaling. Dirchelet calibration improves ECE from 0.1437 to 0.0410 and ACE from 0.1437 to 0.0973. We also compare our results with baseline agents fine tuned with Focal Loss and Label Smoothing.

\noindent \textbf{Debate with Calibrated Agents}
From Fig.~\ref{fig:combined-calibration}, we see that in the Agentic debate framework when the agents are finetuned with \cal loss, it leads to better calibration. ECE reduces from 0.375 to 0.098, MCE from 0.818 to 0.267 and ACE from 0.207 to 0.095 on VQARad dataset. On ScienceQA dataset ECE reduces from 0.396 to 0.055, ACE reduces from 0.372 to 0.110 and MCE from 0.449 to  0.331. The significant reduction in both across the VQARad and ScienceQA datasets—relative to using fine-tuning or debate in isolation—indicates that the calibrated-agents debate yields substantially more reliable confidence estimates and, consequently, more trustworthy answers. 

\mypara{Supplemental materials include}
\textbf{(a)} detailed training procedures with hyperparameters ; \textbf{(b)} detailed analysis of SOTA VLM Models and their performance \textbf{(c)} detailed description of metrics  \textbf{(d)} additional ablations on our debate framework and \textbf{(e)} Broader impact of the research. 

\section{Conclusion}

In this work, we propose AlignVQA, a novel method to improve confidence calibration in Visual Question Answering (VQA) using a multi-agent debate framework. AlignVQA tackles overconfident miscalibration in state-of-the-art VQA models, enhancing reliability in high-stakes domains. Our key contribution is the Agentic Debate Framework with a differentiable calibration-aware loss (AlignCal) that effectively reduces miscalibration. Overall, AlignVQA delivers better-calibrated, more reliable answers. 


\bibliography{aaai2026}
\clearpage
\appendix

\section{Appendix}


We supplement the main text with additional results that comprise the following details.
\begin{itemize}

\item \textbf{Pseudocode:} We present an algorithm that describes the step-by-step procedure for fine-tuning vision-language models on the VQA task, incorporating our proposed calibration loss (\cal loss)

	\item \textbf{Architecture/finetuning procedure}: We briefly describe the architecture of the VLMs used. Additionally, we include the fine-tuning procedure and implementation details.

    \item \textbf{Theoretical insight:} We provide a detailed formulation and a theoretical justification for why the proposed loss function minimizes the expected calibration error.

\item \textbf{Additional Results} Main text comprised of calibration comparison in Base VLM Gemma 3 4B model, FL finetuned Gemma 3 4B model, FL + \cal finetuned Gemma 3 4B model and Agentic Framework results. Supplementary includes calibration performance of different other SOTA VLM Models. We also illustrate here the calibration comparison between FL and FL + \cal fine-tuned LLAVA Onevision model on ScienceQA and VQARad dataset.

\item \textbf{Ablation Studies} Ablation studies on number of agents, debate rounds and \cal loss

\item \textbf{Broader Impact}

\item \textbf{Reproducibility Checklist} 

\end{itemize}

\section{Pseudocode}

We provide the pseudo-code of our finetuning procedure in Alg.~\ref{alg:vqa-fl-ubcef}.

\begin{algorithm}[htb]
\caption{Fine‐tuning VQA Model with Focal + \cal Loss}
\label{alg:vqa-fl-ubcef}
\textbf{Input}: $\{(I_i, Q_i, y_i)\}_{i=1}^N$ (images, questions, ground‐truth answers), Pre-trained VLM $\mathcal{M}$ with parameters $\phi$

\textbf{Parameter}: learning rate $\eta$, epochs $E$, LoRA parameters $\theta$,  batch size $B$, focal parameter $\gamma$, \cal weight $\alpha$

\textbf{Output}: Calibrated fine-tuned VLM $\mathcal{M}'$

\begin{algorithmic}[1]
  \STATE Initialize LoRA parameters $\theta$, and freeze $\phi$
  \FOR{$\text{epoch} = 1$ to $E$}
    \FOR{each batch $B \subseteq \mathcal{D}$}
      \STATE Compute forward pass to get $P_{\phi+\theta}(w_i \mid w_{0:i-1})$ 
      \STATE Compute \cal +FL loss  $\mathcal{L}_{\phi+\theta}^{\text{\cal}+FL}$ 
      \STATE Compute gradients of loss function w.r.t. $\theta$: $\nabla_{\theta} \mathcal{L}_{\phi+\theta}^{\text{\cal}+FL}$
      \STATE Update LoRA parameters: $\theta \gets \theta - \eta \cdot \nabla_{\theta} \mathcal{L}_{\phi+\theta}^{\text{\cal}+FL}$
    \ENDFOR
  \ENDFOR
   \STATE $\theta^* \gets \theta$
    \STATE $\mathcal{M}' \gets \mathcal{M}$ with updated LoRA parameters $\theta^*$
    \RETURN $\mathcal{M}'$
\end{algorithmic}
\end{algorithm}

\subsection{Architectures}

We describe below the architectures that we experimented with.

\begin{figure*}[p]
  \centering
  \begin{subfigure}[b]{0.24\textwidth}
    \centering
    \includegraphics[width=\linewidth]{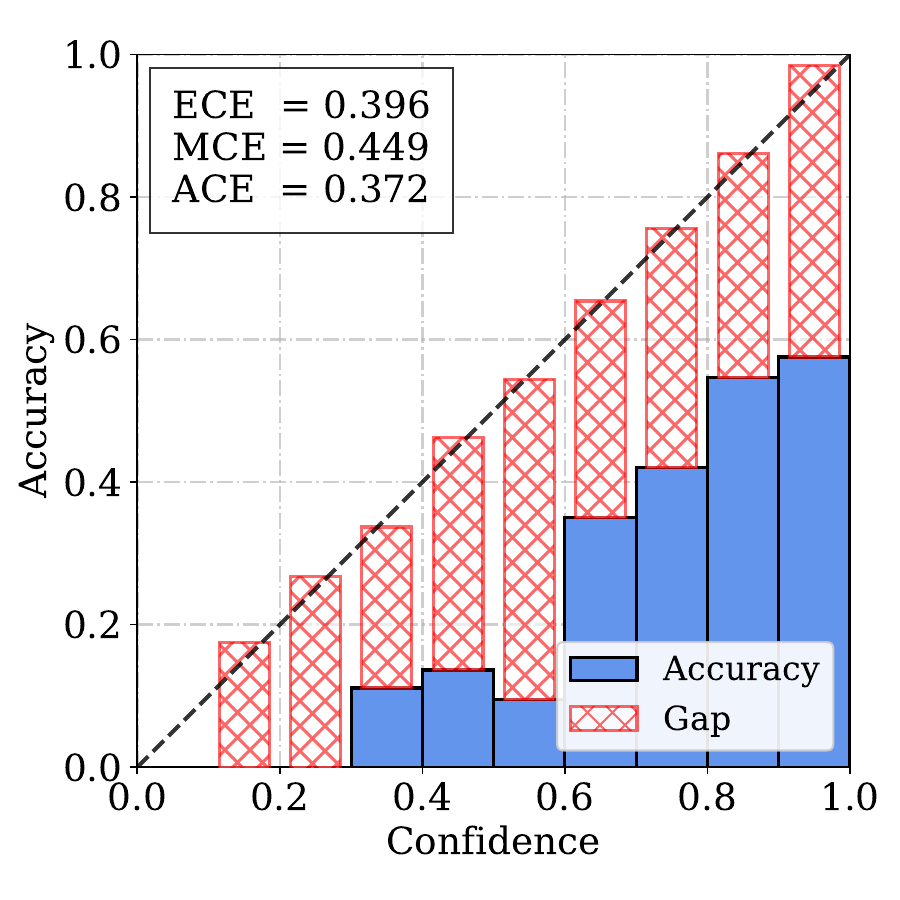}
    \caption{Gemma‐3‐4B}
    \label{fig:all_calib_a}
  \end{subfigure}\hfill
  \begin{subfigure}[b]{0.24\textwidth}
    \centering
    \includegraphics[width=\linewidth]{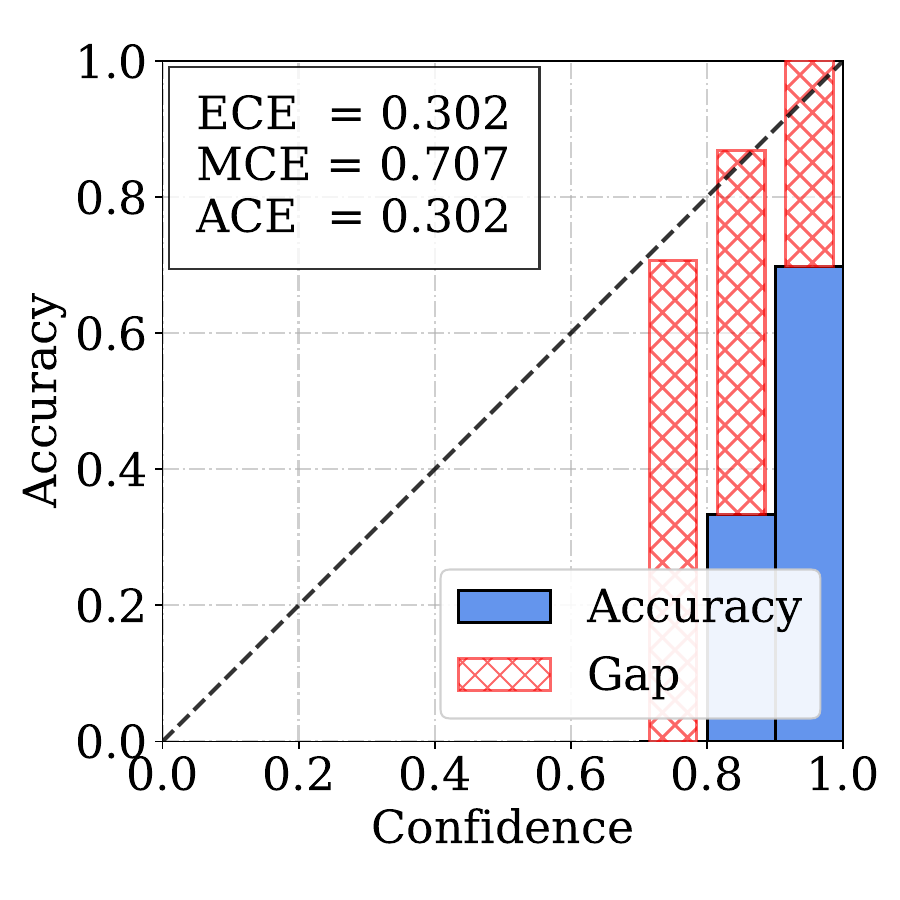}
    \caption{Qwen2.5‐VL‐3B‐Instruct}
    \label{fig:all_calib_b}
  \end{subfigure}\hfill
  \begin{subfigure}[b]{0.24\textwidth}
    \centering
    \includegraphics[width=\linewidth]{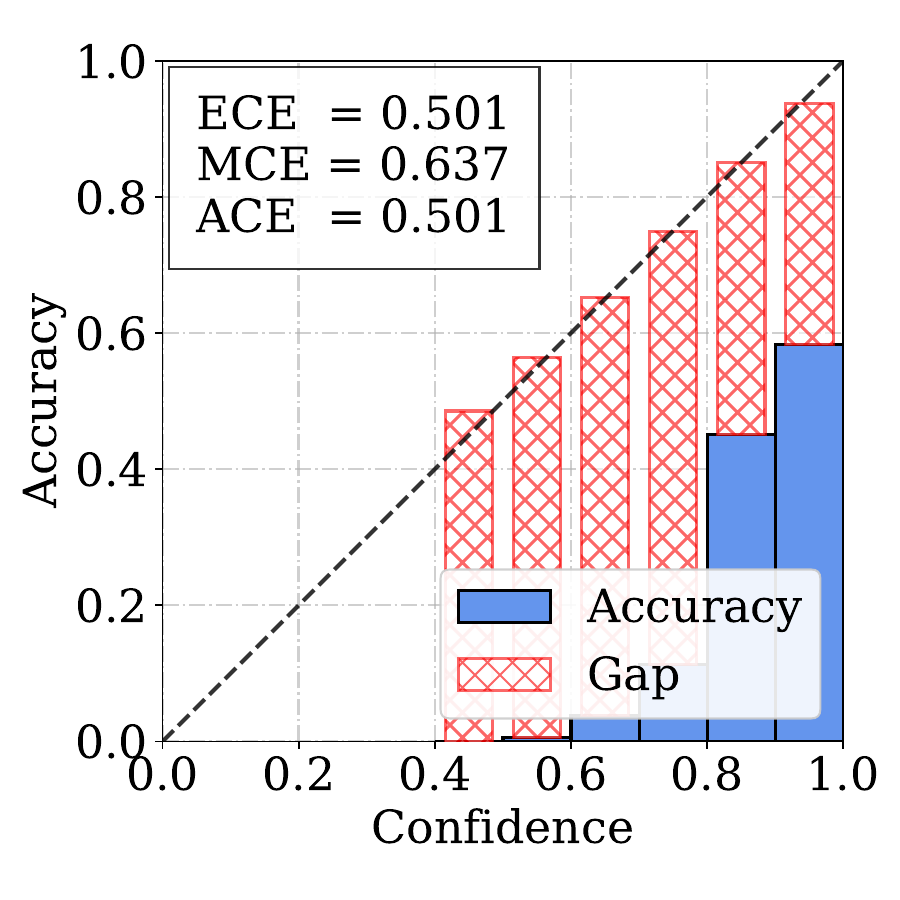}
    \caption{LLAVA‐OneVision}
    \label{fig:all_calib_c}
  \end{subfigure}\hfill
  \begin{subfigure}[b]{0.24\textwidth}
    \centering
    \includegraphics[width=\linewidth]{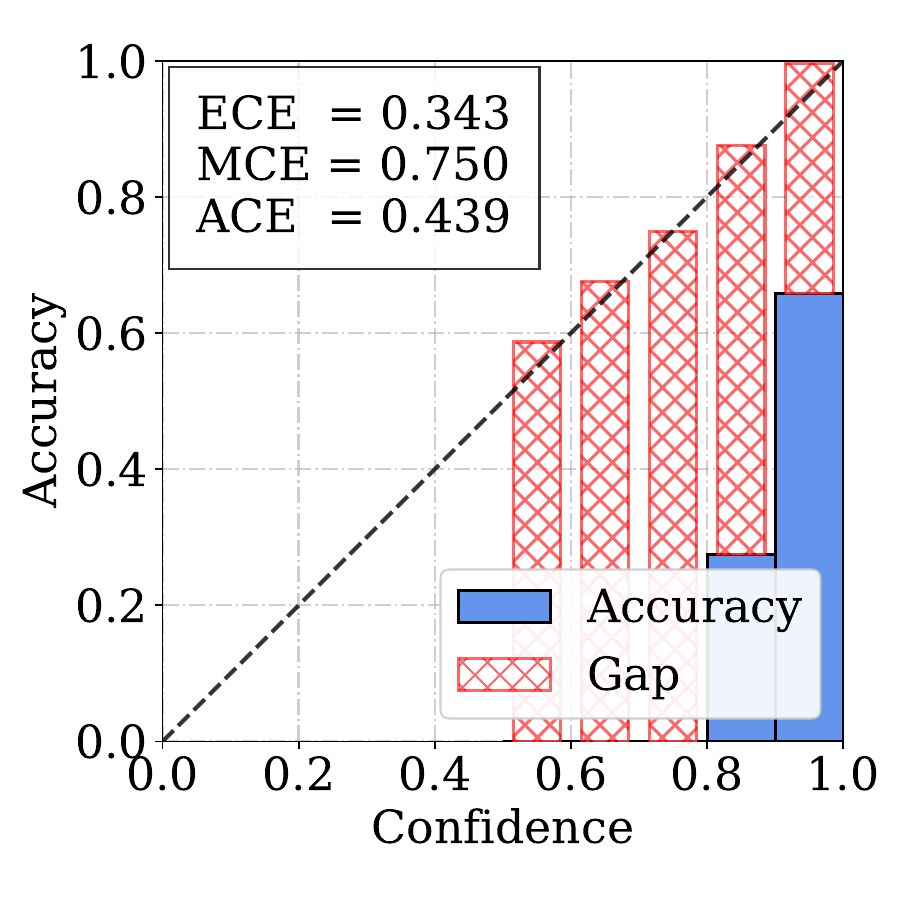}
    \caption{Granite‐vision‐3.3‐2B}
    \label{fig:all_calib_d}
  \end{subfigure}

  \vspace{4pt}

  \begin{subfigure}[b]{0.24\textwidth}
    \centering
    \includegraphics[width=\linewidth]{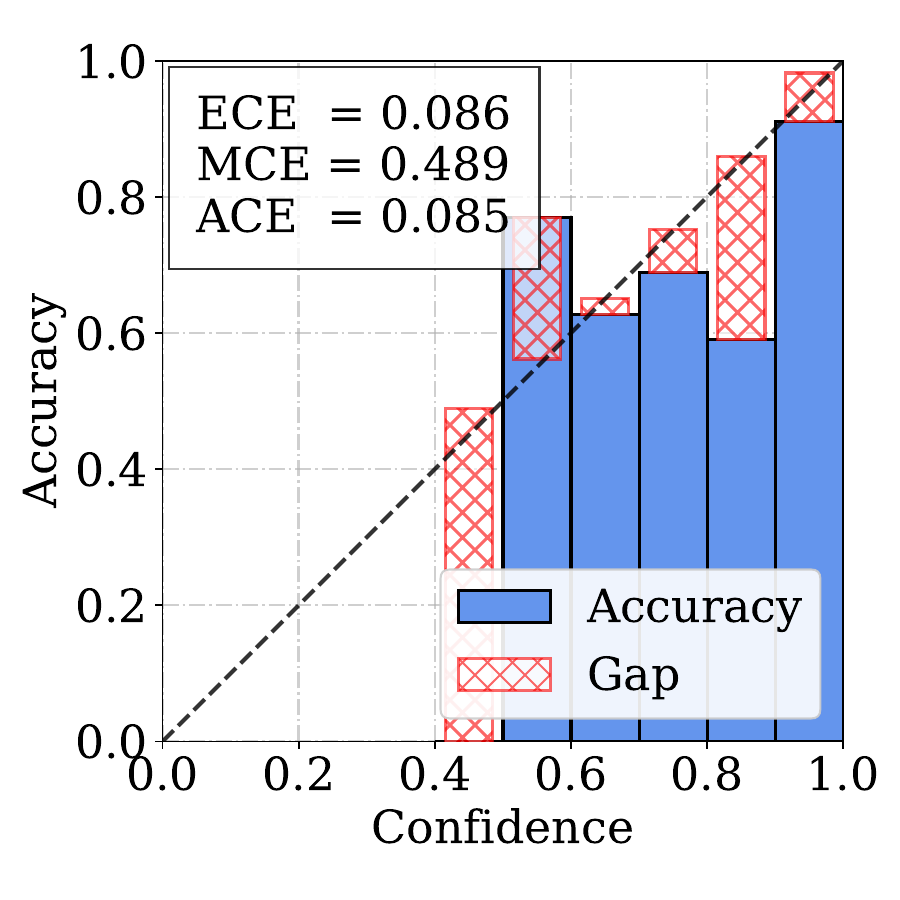}
    \caption{Phi‐4‐Multimodal‐Instruct}
    \label{fig:all_calib_e}
  \end{subfigure}\hfill
  \begin{subfigure}[b]{0.24\textwidth}
    \centering
    \includegraphics[width=\linewidth]{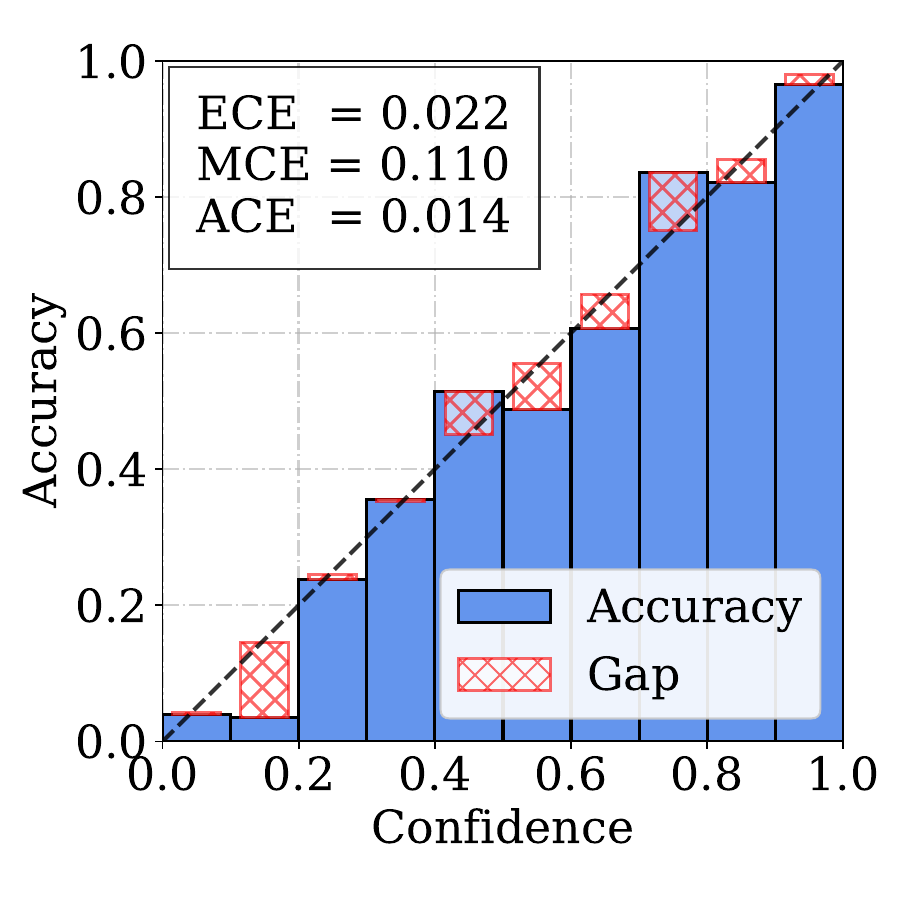}
    \caption{InternVL‐4B}
    \label{fig:all_calib_f}
  \end{subfigure}\hfill
  \begin{subfigure}[b]{0.24\textwidth}
    \centering
    \includegraphics[width=\linewidth]{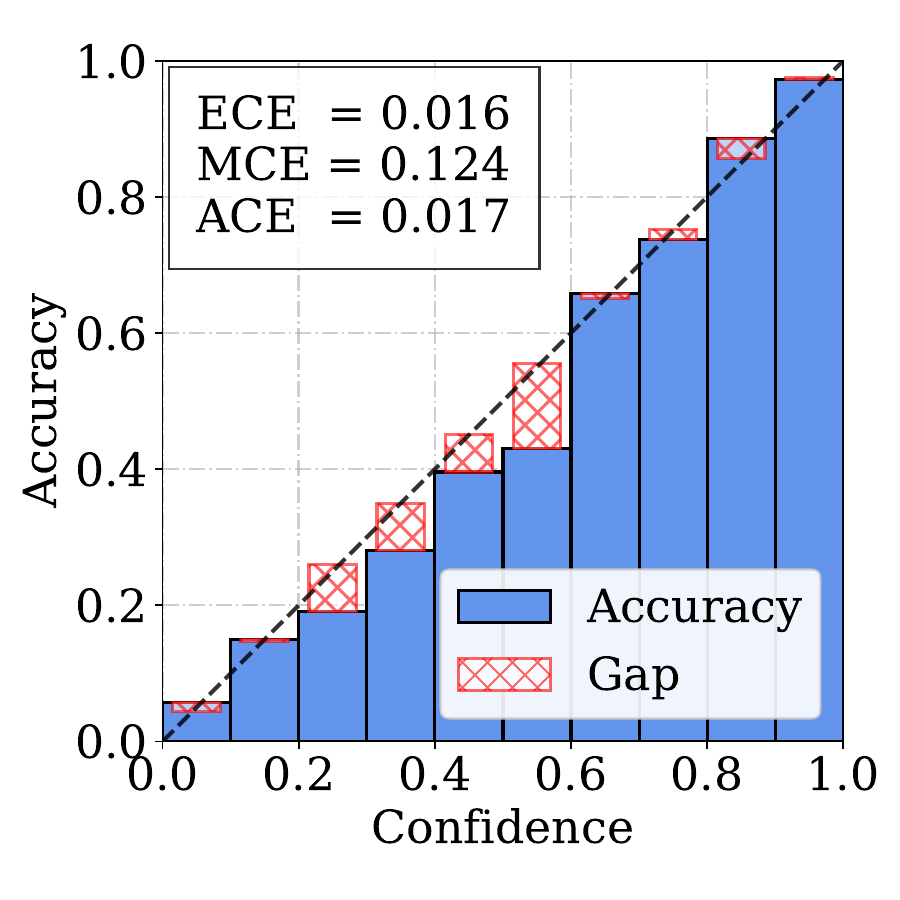}
    \caption{Ristretto‐3B}
    \label{fig:all_calib_g}
  \end{subfigure}\hfill
  \begin{subfigure}[b]{0.24\textwidth}
    \centering
    \includegraphics[width=\linewidth]{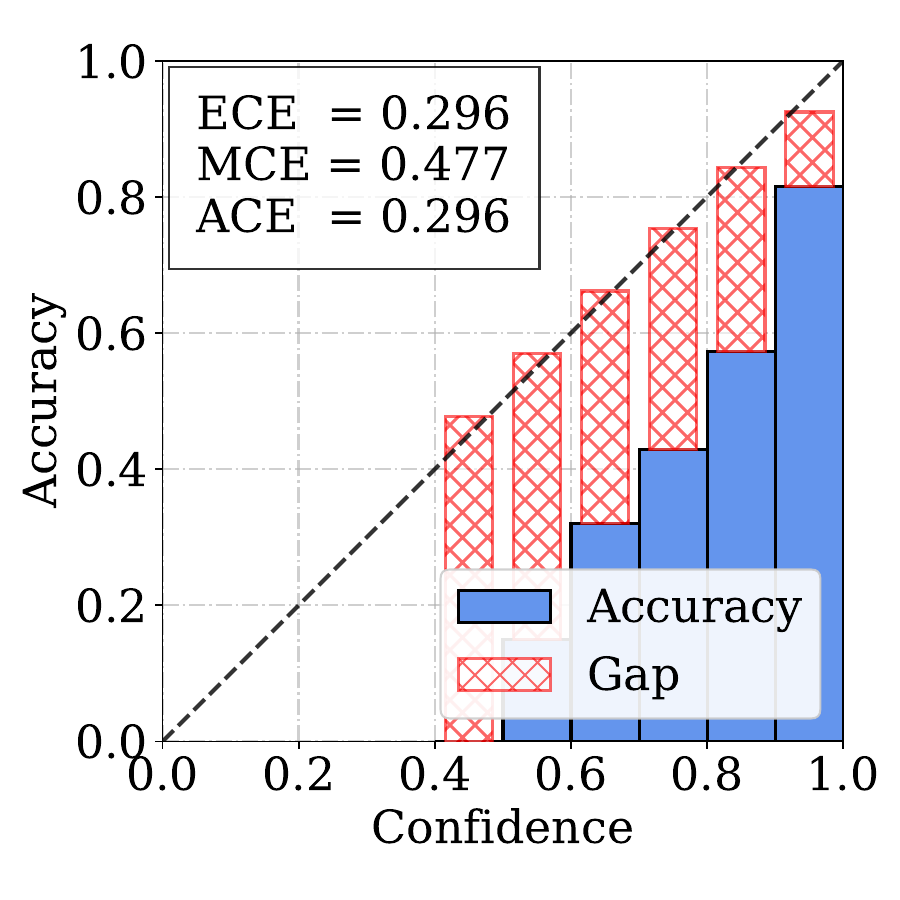}
    \caption{SmolVLM}
    \label{fig:all_calib_h}
  \end{subfigure}

  \vspace{4pt}

  \centering
  \begin{subfigure}[b]{0.24\textwidth}
    \centering
    \includegraphics[width=\linewidth]{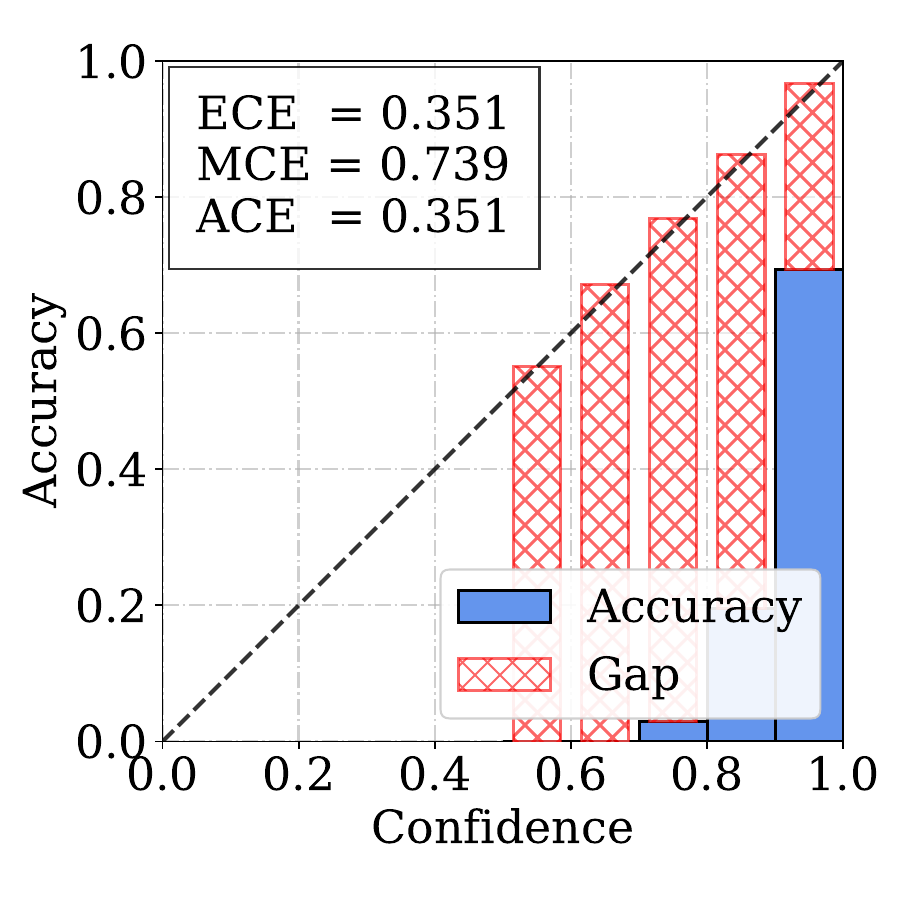}
    \caption{Ovis2-4B}
    \label{fig:all_calib_i}
  \end{subfigure}\hspace{6pt}
  \begin{subfigure}[b]{0.24\textwidth}
    \centering
    \includegraphics[width=\linewidth]{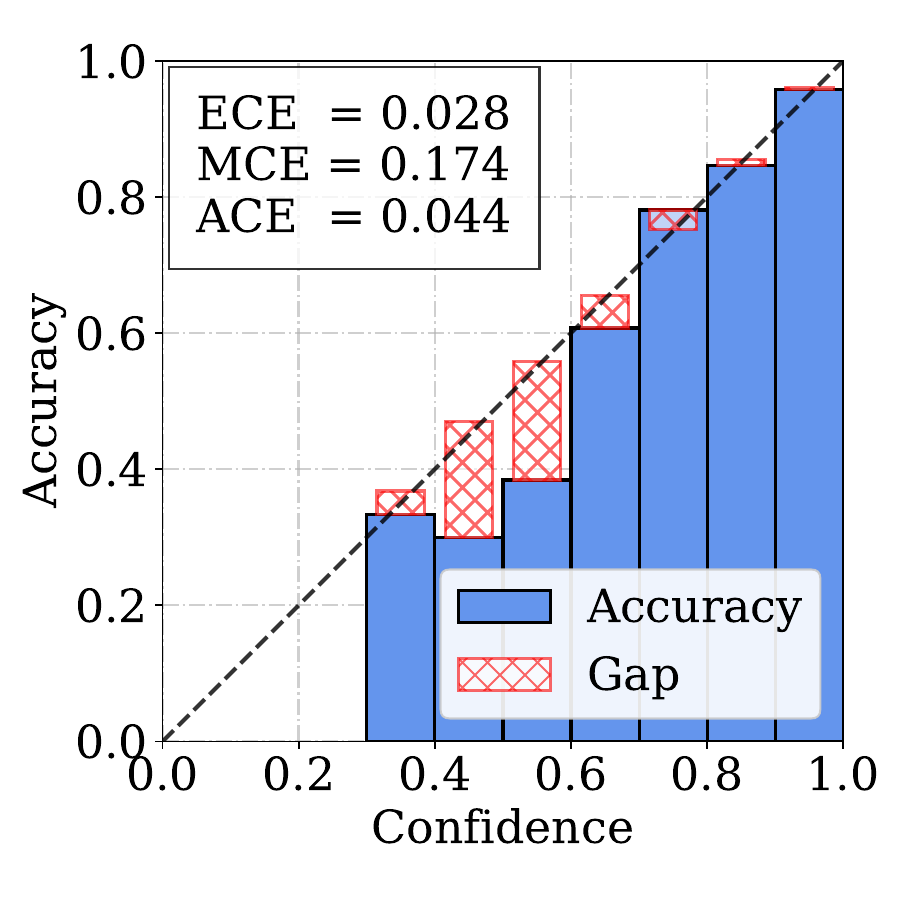}
    \caption{DeepSeekVL2-Tiny}
    \label{fig:all_calib_j}
  \end{subfigure}

  \caption{Reliability plots for base VLMs on ScienceQA for (a) Gemma‐3‐4B, (b) Qwen2.5‐VL‐3B‐Instruct, (c) LLAVA‐OneVision, (d) Granite‐vision‐3.3‐2B, (e) Phi‐4‐Multimodal‐Instruct, (f) InternVL‐4B, (g) Ristretto‐3B, (h) SmolVLM, (i) Ovis2-4B, and (j) DeepSeekVL2-Tiny. We observe that majority of the base VLM models are overconfident except for \ref{fig:all_calib_f}, \ref{fig:all_calib_g} and \ref{fig:all_calib_j}. }
  \label{fig:all_scienceqa}
\end{figure*}

\begin{figure*}[p]
  \centering
  \begin{subfigure}[b]{0.24\textwidth}
    \centering
    \includegraphics[width=\linewidth]{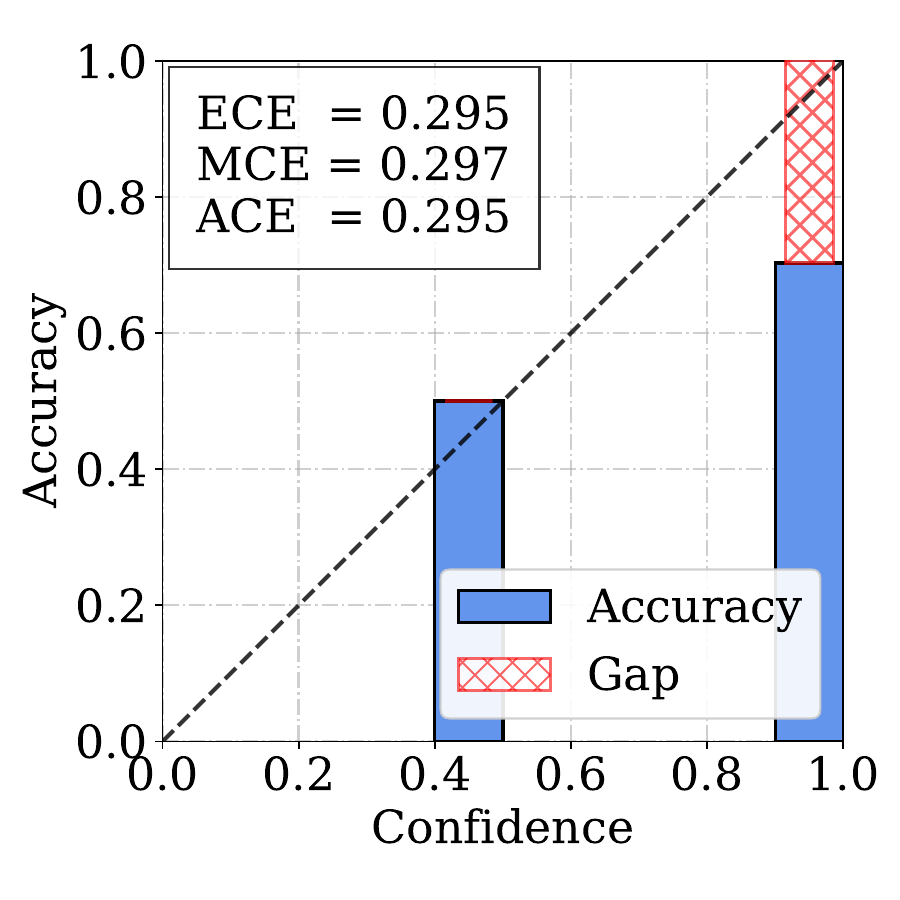}
    \caption{Qwen2.5-VL-3B-Instruct}
    \label{fig:rad_a}
  \end{subfigure}\hfill
  \begin{subfigure}[b]{0.24\textwidth}
    \centering
    \includegraphics[width=\linewidth]{sections/gemmavqarad.pdf}
    \caption{Gemma-3-4B}
    \label{fig:rad_b}
  \end{subfigure}\hfill
  \begin{subfigure}[b]{0.24\textwidth}
    \centering
    \includegraphics[width=\linewidth]{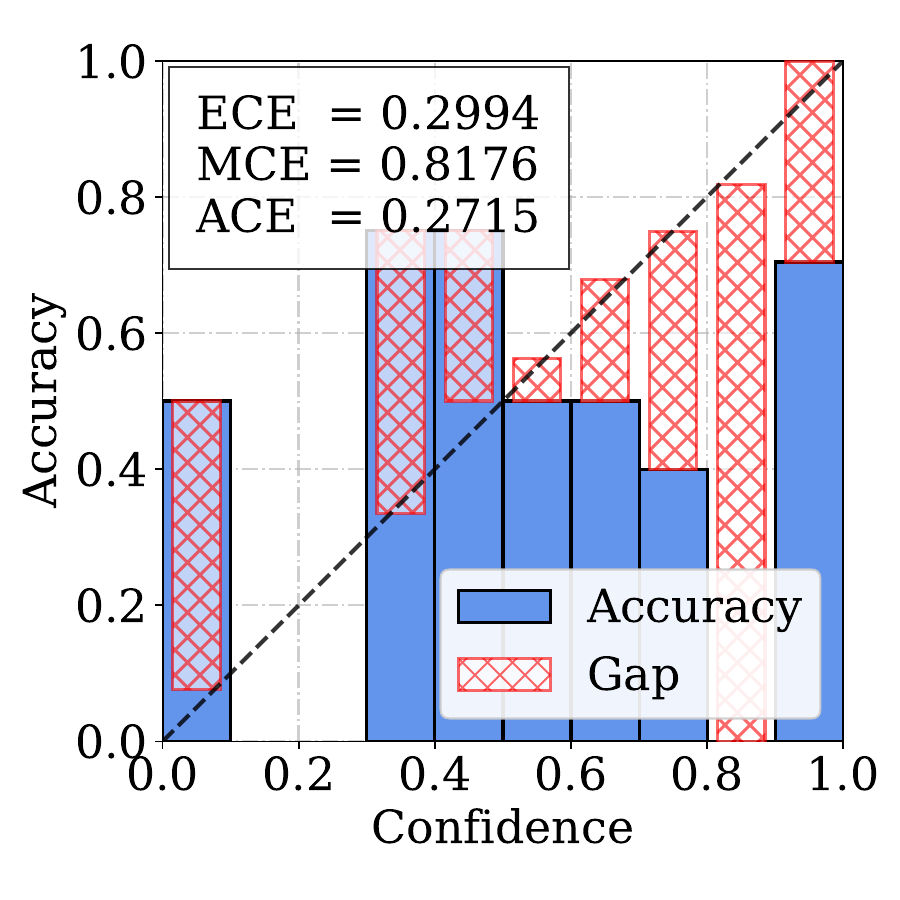}
    \caption{LLAVA-OneVision}
    \label{fig:rad_c}
  \end{subfigure}\hfill
  \begin{subfigure}[b]{0.24\textwidth}
    \centering
    \includegraphics[width=\linewidth]{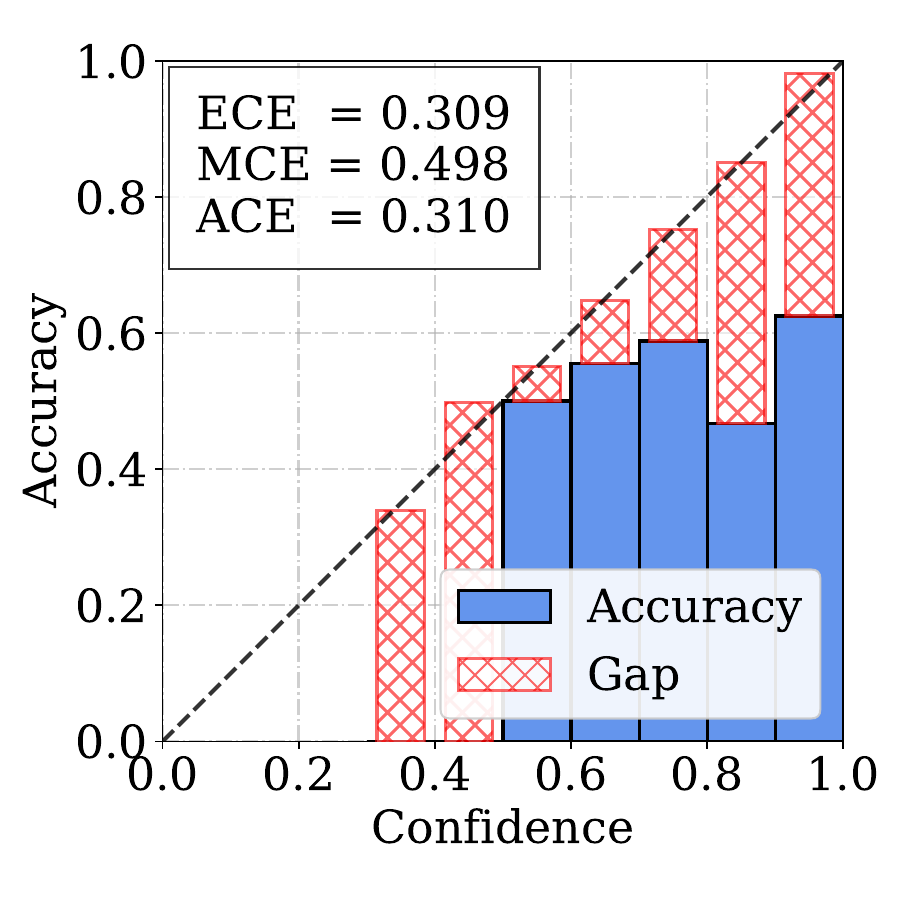}
    \caption{Granite-vision-3.3-2B}
    \label{fig:rad_d}
  \end{subfigure}

  \vspace{4pt}

  \begin{subfigure}[b]{0.24\textwidth}
    \centering
    \includegraphics[width=\linewidth]{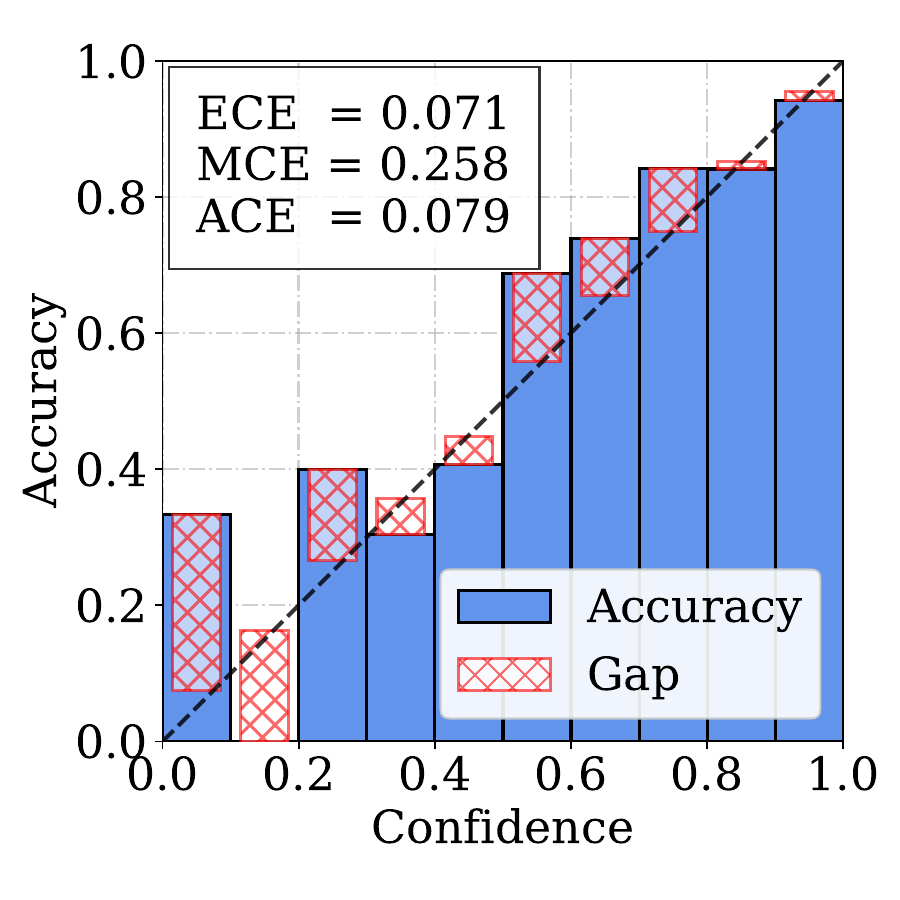}
    \caption{Ristretto-3B}
    \label{fig:rad_e}
  \end{subfigure}\hfill
  \begin{subfigure}[b]{0.24\textwidth}
    \centering
    \includegraphics[width=\linewidth]{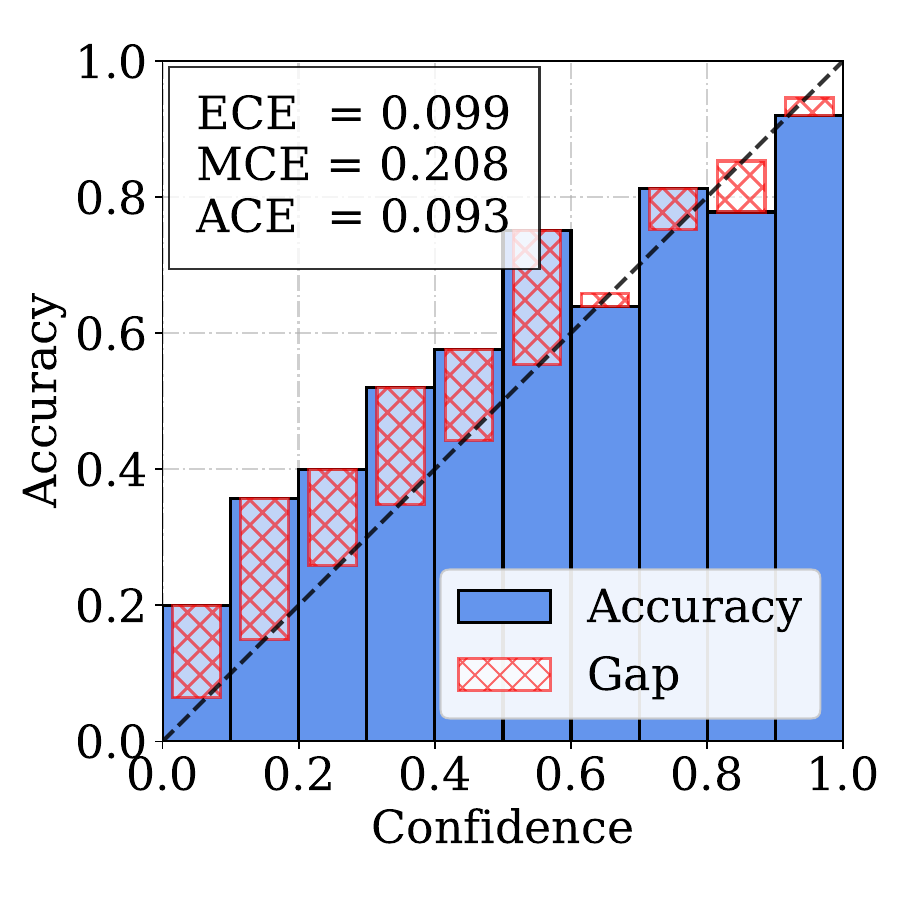}
    \caption{InternVL-4B}
    \label{fig:rad_f}
  \end{subfigure}\hfill
  \begin{subfigure}[b]{0.24\textwidth}
    \centering
    \includegraphics[width=\linewidth]{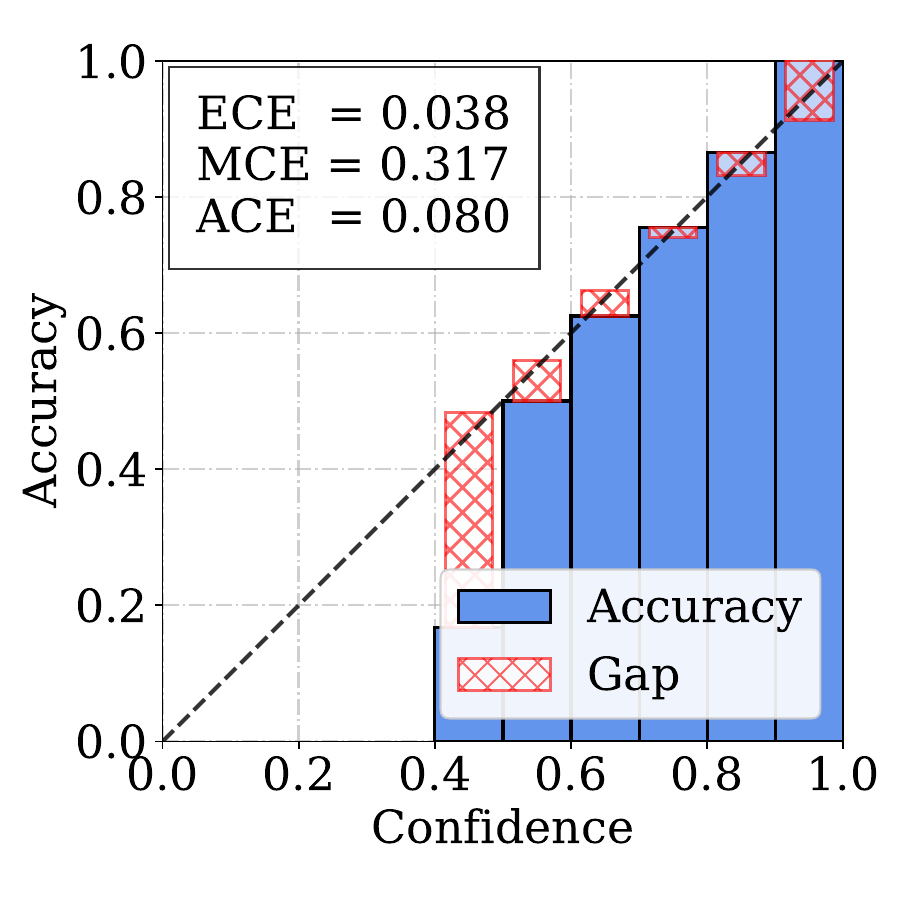}
    \caption{Ovis2-4B}
    \label{fig:rad_g}
  \end{subfigure}\hfill
  \begin{subfigure}[b]{0.24\textwidth}
    \centering
    \includegraphics[width=\linewidth]{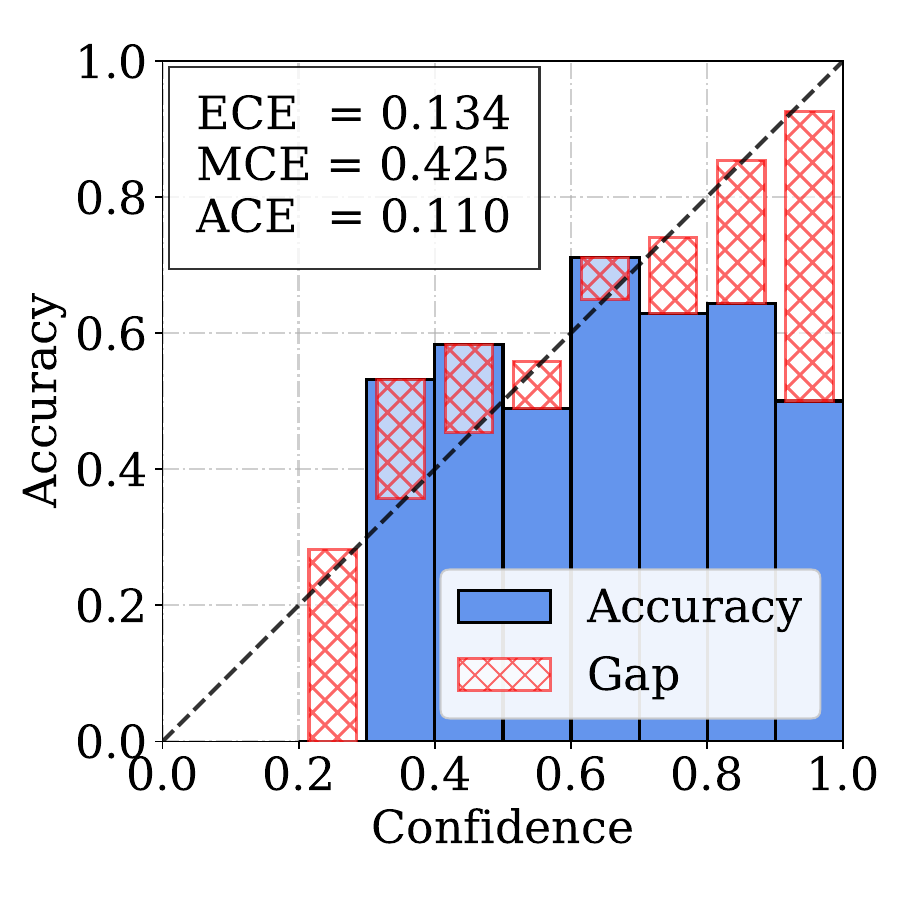}
    \caption{Phi-4-Multimodal-Instruct}
    \label{fig:rad_h}
  \end{subfigure}

  \vspace{4pt}

  \begin{subfigure}[b]{0.24\textwidth}
    \centering
    \includegraphics[width=\linewidth]{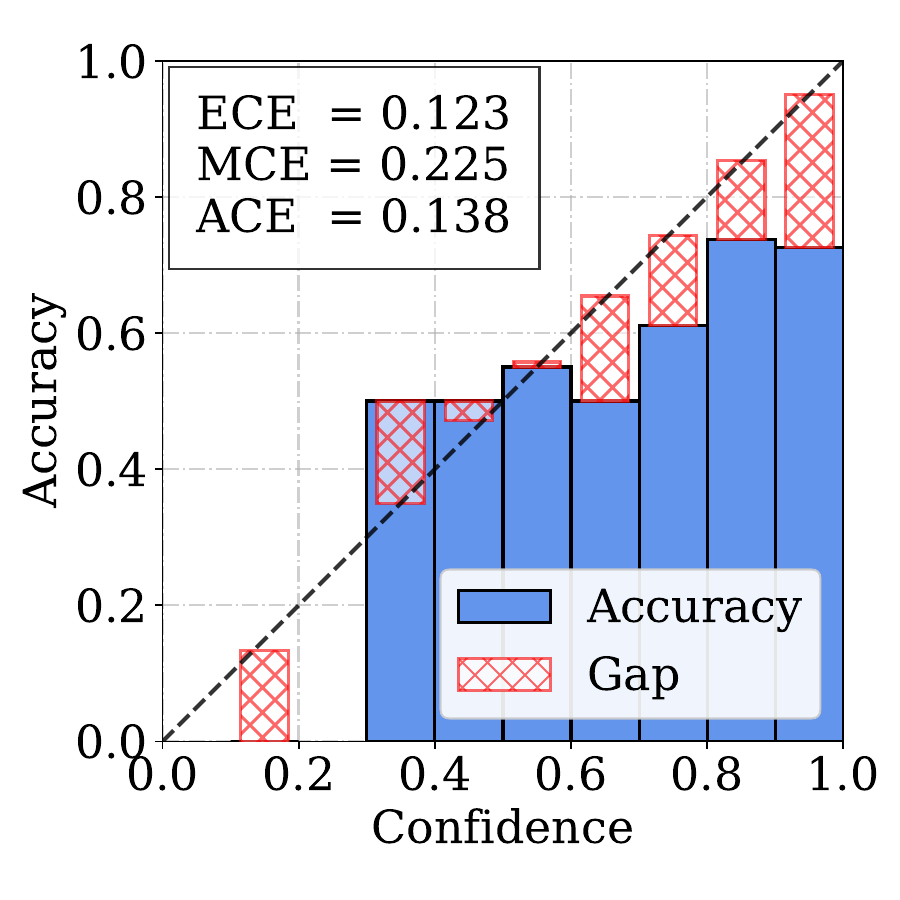}
    \caption{SmolVLM}
    \label{fig:rad_i}
  \end{subfigure}\hspace{6pt}
  \begin{subfigure}[b]{0.24\textwidth}
    \centering
    \includegraphics[width=\linewidth]{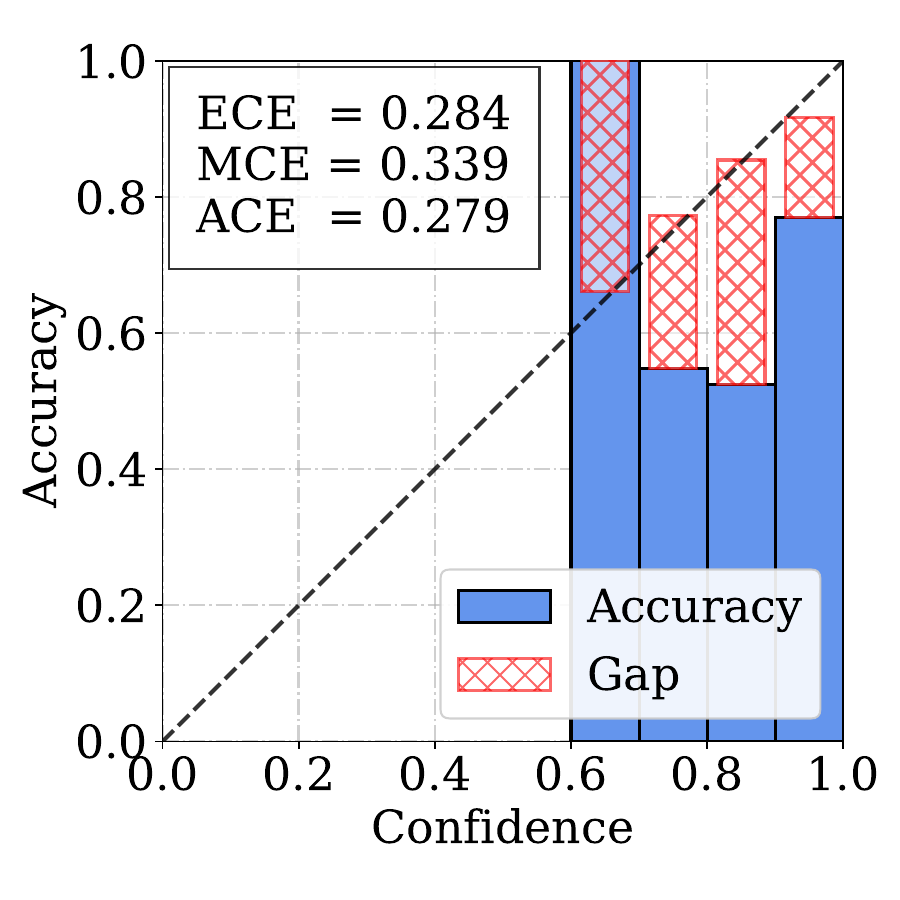}
    \caption{DeepSeek VL2-Tiny}
    \label{fig:rad_j}
  \end{subfigure}

  \caption{Reliability plots for base VLMs on VQARad for (a) Qwen2.5-VL-3B-Instruct, (b) Gemma-3-4B, (c) LLAVA-OneVision, (d) Granite-vision-3.3-2B, (e) Ristretto-3B, (f) InternVL-4B, (g) Ovis2-4B, (h) Phi-4-Multimodal-Instruct, (i) SmolVLM, and (j) DeepSeek VL2-Tiny. We observe that majority of the base VLM models are either overconfident or underconfident except for \ref{fig:rad_f} and \ref{fig:rad_g} on the VQARad dataset.}
  \label{fig:all_vqarad}
\end{figure*}

\begin{enumerate}
    \item \textbf{LLAVA-OneVision \cite{li2024llava}} is a Vision-Language Model capable of generating text conditioned on one or more images or videos. It also allows for strong transfer learning across different scenarios (single image, multi image, and video) enhancing the capabilities of the contemporary VLM scene. It combines a SigLIP vision encoder with a Qwen2 language backbone for effective results.
    \item \textbf{Gemma 3 4B \cite{team2025gemma}} is part of Google DeepMind's Gemma3 family. It utilizes a SigLIP vision encoder with Gemma 4B as the language backbone. It is compact enough for broader deployment and is well-suited for a variety of text generation and image understanding tasks like question answering, summarization, and reasoning.
    \item \textbf{Qwen/Qwen2.5-VL-3B-Instruct\cite{bai2025qwen2}} is a 3B-parameter Vision-Language Instruct model from the Qwen/VL series. For the vision encoder, it implements window attention into the ViT, which is then further optimized with SwiGLU and RMSNorm to align it with the Qwen2.5 VL architecture. The language backbone comprises of 3B instruction-tuned variant of the Qwen2.5 LLM. It's designed for generating structured outputs using it's proficient reasoning and analysis skills.
    \item \textbf{Phi-4-multimodal-instruct\cite{abouelenin2025phi}} is a lightweight open-source multimodal transformer model with 5.6B parameters. It processes text, image, and audio inputs to produce enhanced text outputs, which is possible due to supervised fine-tuning, fine-tuning on human preferences, and RLHF. The model architecture is composed of multiple advanced image encoders and the pretrained Phi-4-Mini-Instruct as the backbone language model.
    
\end{enumerate}
The calibration results for different VLM models mentioned in Table \ref{tab:sota_combined_vlm} are illustrated in Figures \ref{fig:all_scienceqa} and \ref{fig:all_vqarad}.
\section{Finetuning Procedure and Implementation}
In our agentic framework, we employ Qwen2.5-VL-3B-Instruct \cite{bai2025qwen2}, Llava-OneVision \cite{li2024llava}, Gemma-3-4B \cite{team2025gemma}, and Phi-4-Multimodal-Instruct \cite{abouelenin2025phi}, each quantized to 4-bit precision using the BitsAndBytes library. We adopt standard chain-of-thought prompting and a knowledge-based agent architecture.
\\
All experiments were conducted on an NVIDIA A100 40 GB GPU. We perform LoRA fine-tuning with rank 8, a scaling factor of 8, and a dropout rate of 0.05. For attention-only adapters, LoRA modules are injected into the \texttt{q\_proj} and \texttt{v\_proj} layers. Calibration fine-tuning is run for 6 epochs on the VQARad dataset and 10 epochs on the ScienceQA dataset. We use a batch size of 2 and optimize with the fused AdamW optimizer, starting from a learning rate of $2\times10^{-4}$. The ScienceQA training set comprises 12 000 examples, while VQARad contains 1 793 examples. For the calibration loss, we set $\lambda=2$.
\\
\textbf{PostHoc Calibration}
For comparison with post-hoc techniques, we set aside 10\% of the training data as a hold out set to conduct post-hoc calibration. For Temperature Scaling (TS), we perform a grid search between the range of 0 to 10 with a step-size of 0.1 to find the optimal temperature value that gives the least NLL on the hold-out set. In Dirichlet Calibration (DC), we append a single‐layer neural network to the DNN’s output and impose ODIR \cite{kull2019beyond} regularization on that layer’s weights.
\\
\textbf{Calibration via finetuning}
We also compared FL + \cal calibration results with other train-time calibration results.
\paragraph{Label Smoothing (LS) \cite{szegedy2016rethinking}}
In LS the hard one‐hot targets are replaced with a smoothed distribution \(q_i\), where for each sample \(i\) and class \(j\):
\[
q_{i,j} \;=\;
\begin{cases}
1 - \alpha\,, & j = y_i,\\
\dfrac{\alpha}{K - 1}\,, & j \neq y_i,
\end{cases}
\]
with \(\alpha\) a hyperparameter (\(\alpha=0.1\) in our experiments).  The label-smoothing loss is then
\[
\mathrm{LS}
= - \sum_{i=1}^{N} \sum_{j=1}^{K}
     q_{i,j}\,\log \hat p_{i,j},
\]
where \(\hat p_{i,j}\) denotes the model’s predicted probability for class \(j\) on sample \(i\).

\paragraph{Focal Loss(FL) \cite{mukhoti2020calibrating}}
Focal loss down‐weights well‐classified examples by a factor \((1 - \hat p_{i,y_i})^\gamma\).  Formally,
\[
\mathrm{FL}
= - \sum_{i=1}^{N}
     \bigl(1 - \hat p_{i,y_i}\bigr)^{\gamma}\,
     \log \hat p_{i,y_i},
\]
where \(\gamma\) is the focusing parameter. In the experiments we used \(\gamma\) to be 2.
\section{Additional Results}
\begin{figure*}[ht]
  \centering
  \begin{subfigure}[b]{0.24\textwidth}
    \centering
    \includegraphics[width=\linewidth]{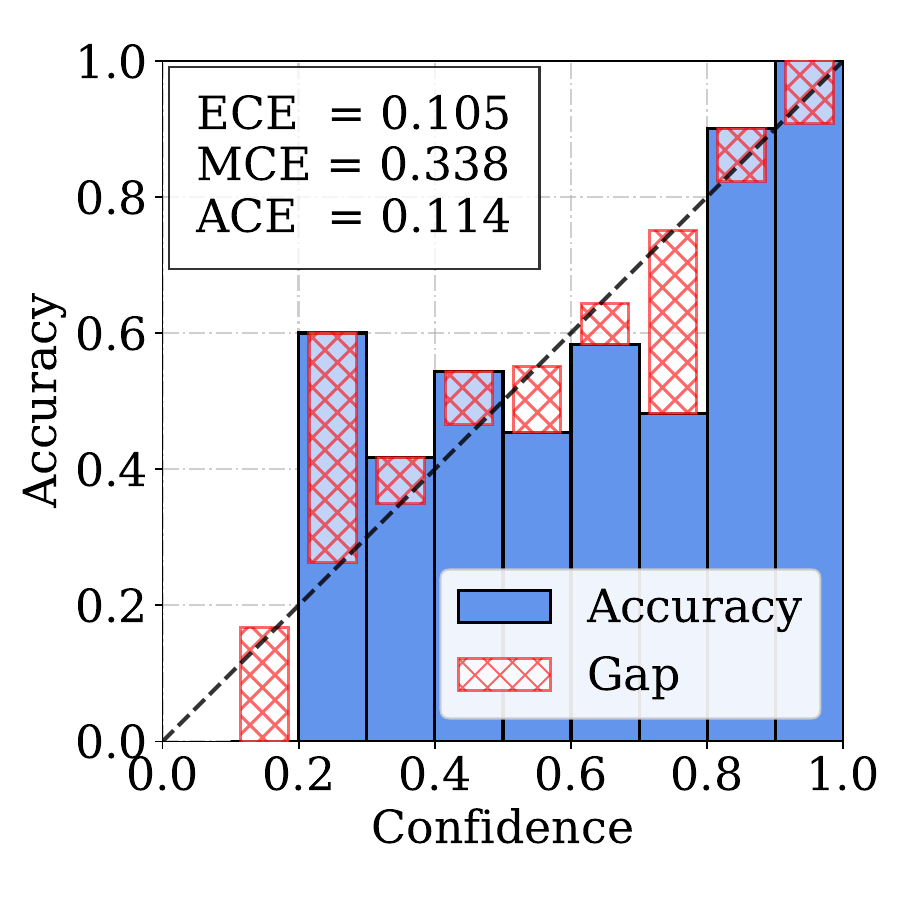}
    \caption{FL}
    \label{fig:flspath}
  \end{subfigure}\hspace{2em}
  \begin{subfigure}[b]{0.24\textwidth}
    \centering
    \includegraphics[width=\linewidth]{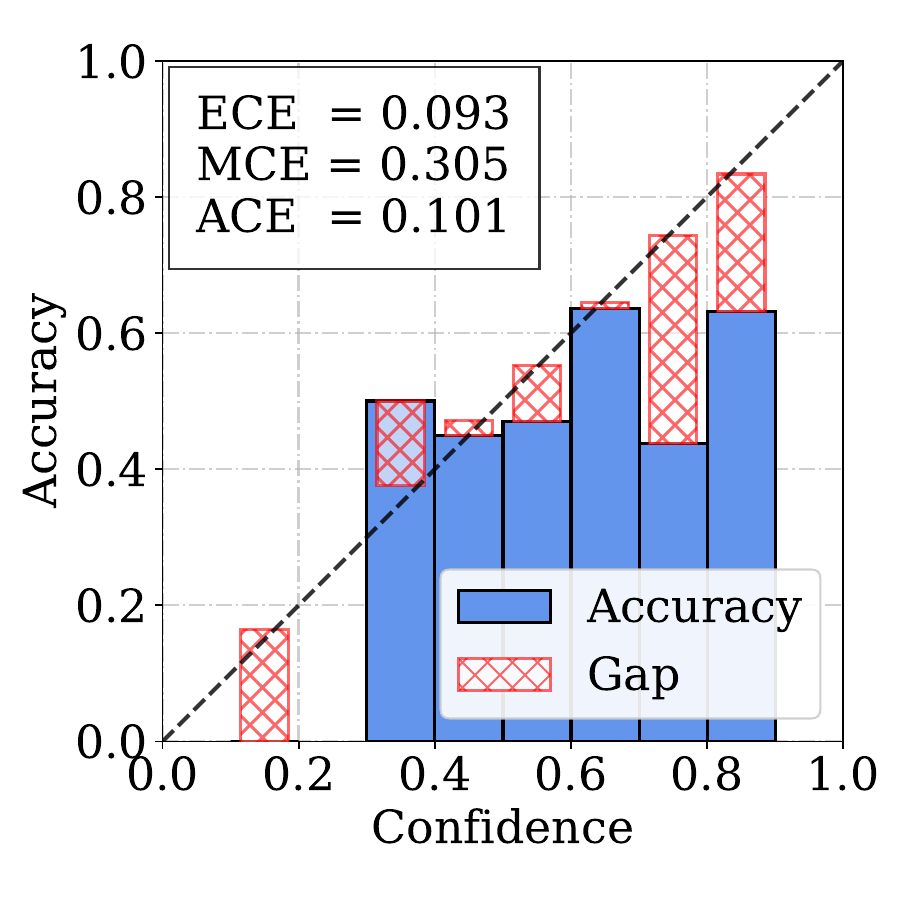}
    \caption{FL + \cal}
    \label{fig:flalignpath}
  \end{subfigure}\\
  \begin{subfigure}[b]{0.24\textwidth}
    \centering
    \includegraphics[width=\linewidth]{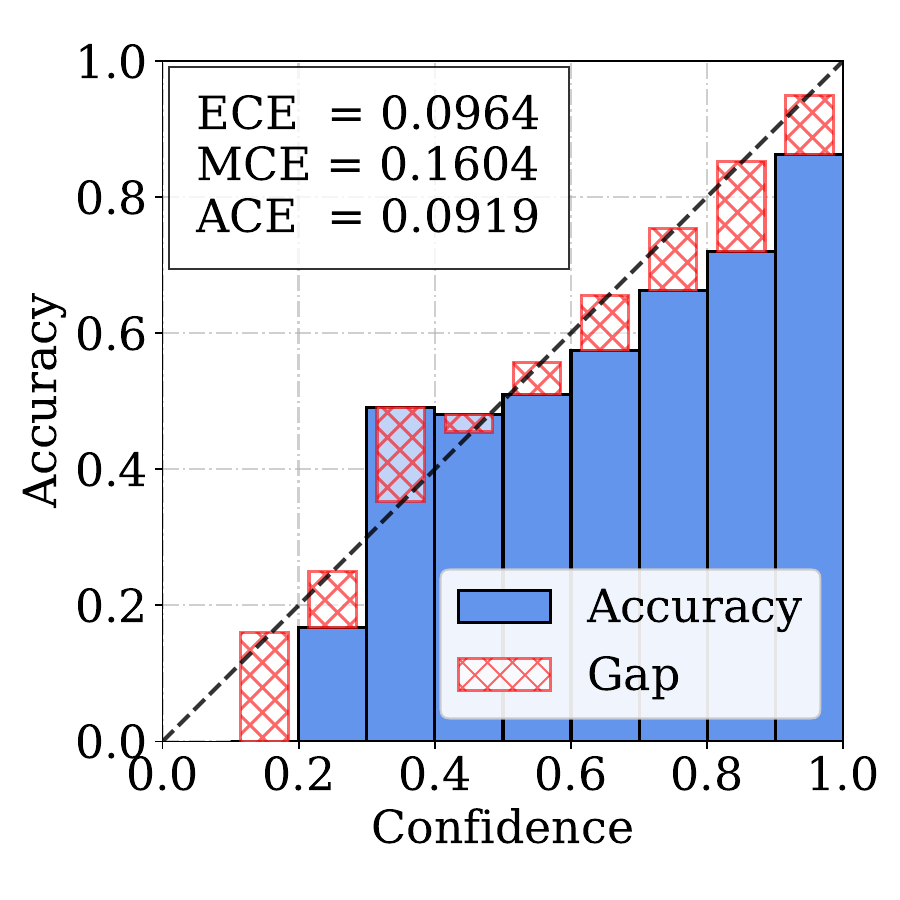}
    \caption{FL}
    \label{fig:flscience}
  \end{subfigure}\hspace{2em}
  \begin{subfigure}[b]{0.24\textwidth}
    \centering
    \includegraphics[width=\linewidth]{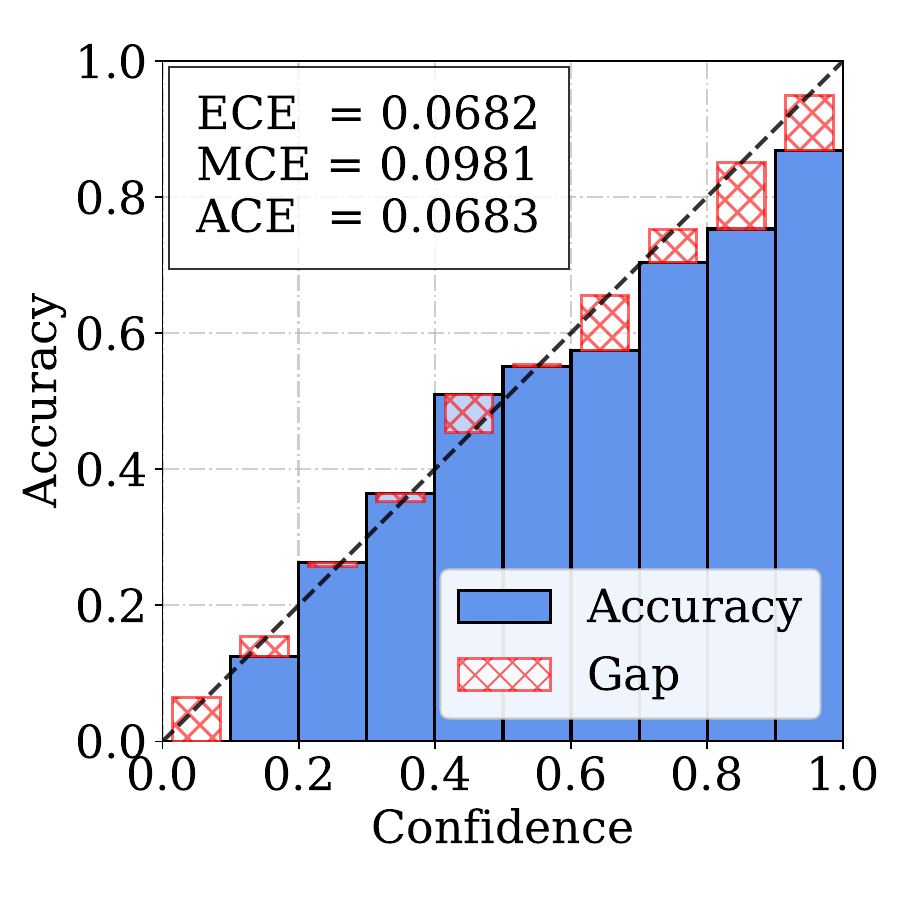}
    \caption{FL + \cal}
    \label{fig:flalignscience}
  \end{subfigure}
  \caption{Reliability plot comparison between (\ref{fig:flspath}) FL and (\ref{fig:flalignpath}) FL + \cal  finetuned LLAVA-OneVision model on VQARad dataset, (\ref{fig:flscience})  FL  and (\ref{fig:flalignscience}) FL + \cal finetuned LLAVA-OneVision model on ScienceQA dataset. 
  Our proposed loss FL + \cal is effective at improving ECE from 10.5\% to 9.3\% on the VQARad dataset and a reduction from 9.64\% to 6.8\% on the ScienceQA dataset. MCE and ACE also follows a similar decreasing trend.}
  \label{fig:all_calib_llava}
\end{figure*}

\begin{figure*}[p]
  \centering
  \begin{subfigure}[b]{0.24\textwidth}
    \centering
    \includegraphics[width=\linewidth]{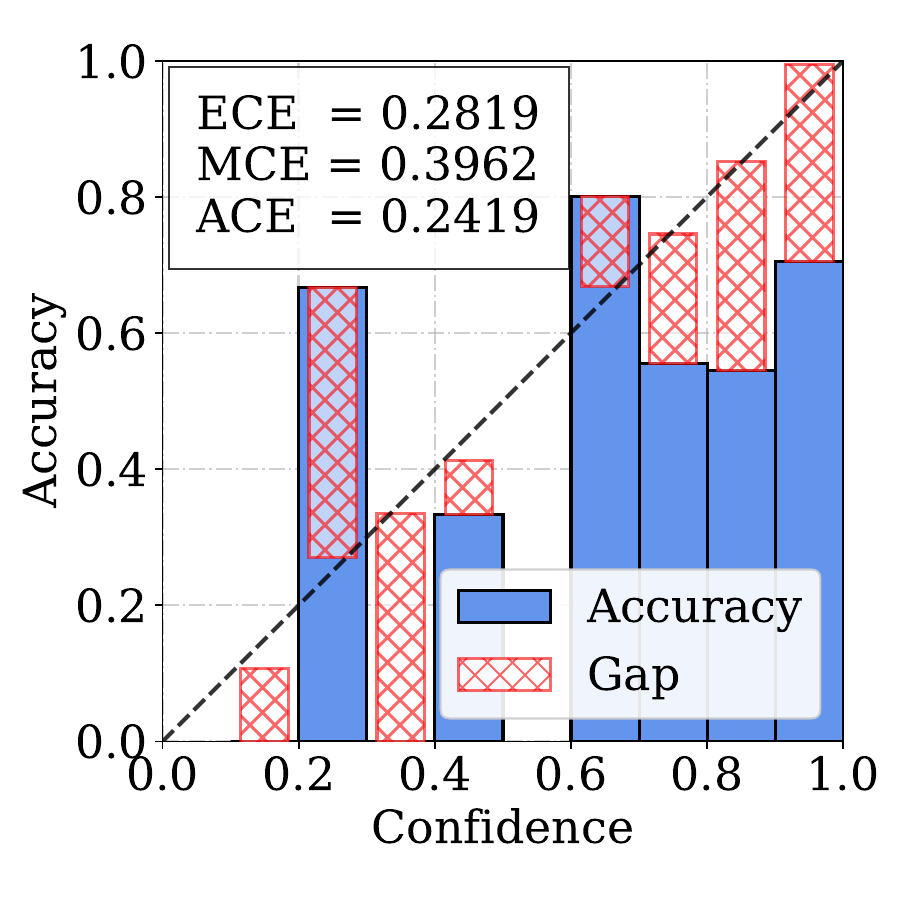}
    \caption{Label Smoothing}
    \label{fig:ls}
  \end{subfigure}\hspace{2em}
  \begin{subfigure}[b]{0.24\textwidth}
    \centering
    \includegraphics[width=\linewidth]{sections/focallossvqaradf.pdf}
    \caption{FL}
    \label{fig:fl}
  \end{subfigure}\hspace{2em}
  \begin{subfigure}[b]{0.24\textwidth}
    \centering
    \includegraphics[width=\linewidth]{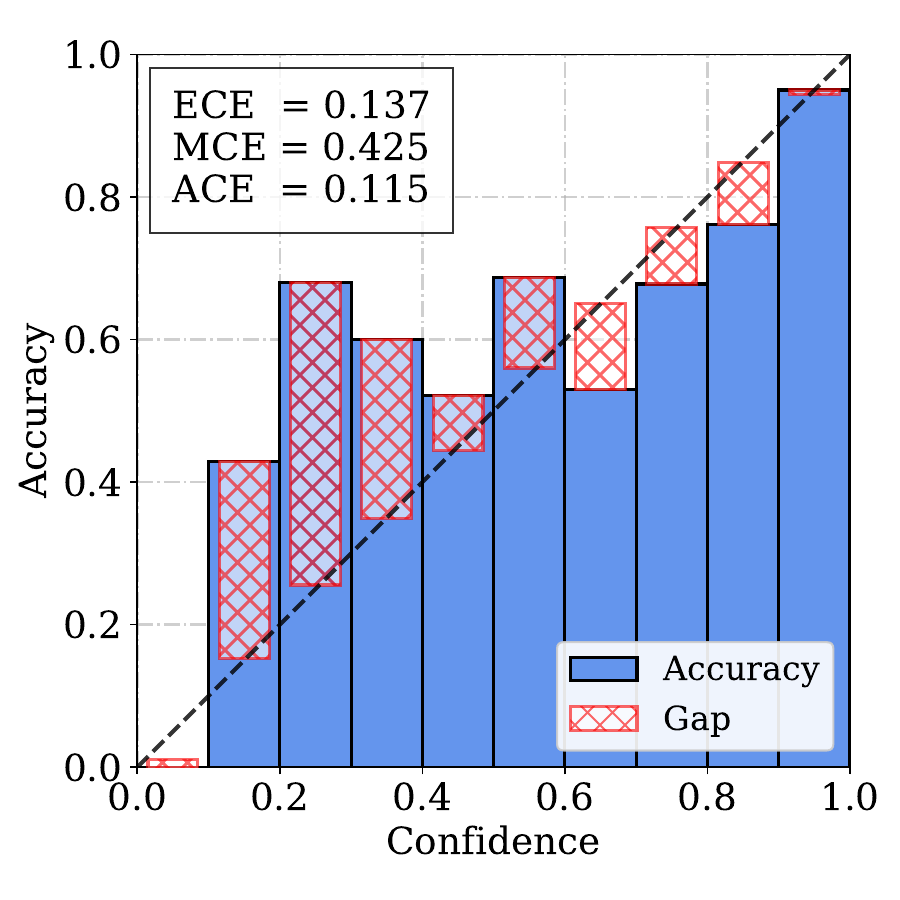}
    \caption{FL + \cal}
    \label{fig:all_calib_flalign}
  \end{subfigure}
    \caption{Reliability plot comparision  between (\ref{fig:ls}) Label Smoothing , (\ref{fig:fl}) FL , (\ref{fig:all_calib_flalign}) FL + \cal finetuned  Gemma 3 4B on VQARad dataset. Our proposed loss FL + \cal is effective at improving ECE from LS to FL + \cal by 15\%. It further shows a improvement of 4\% from FL to FL + \cal on the VQARad dataset. MCE and ACE also follow a similar decreasing trend.}
  \label{fig:all_calib_diff}
\end{figure*}

Figure \ref{fig:all_calib_llava} shows the calibration of the LLAVA‐OneVision model fine‐tuned with focal loss and with focal loss + \cal. On the VQARad dataset, ECE decreases from 0.105 (FL fine‐tuned) to 0.093 (FL + \cal fine‐tuned). ACE decreases from 0.114 to 0.101, and MCE from 0.338 to 0.305. On the ScienceQA dataset, ECE decreases from 0.0964 (focal loss fine‐tuned) to 0.0682 (FL + \cal fine‐tuned). ACE decreases from 0.0919 to 0.0683, and MCE from 0.1604 to 0.0981.\\
Figure \ref{fig:all_calib_diff} shows the calibration comparisons between training‐time calibration methods. The label‐smoothing fine‐tuned model yields an ECE of 0.2819, an MCE of 0.3962, and an ACE of 0.2419. When using the focal‐loss fine‐tuned model, calibration performance improves: ECE decreases to 0.178, MCE to 0.430, and ACE to 0.155. The best calibration performance is achieved by the focal loss + \cal model, with an ECE of 0.137, an MCE of 0.425, and an ACE of 0.115.

\section{Broader Impact}
The proposed approach to improving calibration in agentic Visual Question Answering (VQA) systems has potential implications across both practical and ethical dimensions. By aligning model confidence more closely with correctness, especially in multi-agent architectures, our work enhances the reliability and trustworthiness of AI systems operating under visual uncertainty. This is particularly valuable in high-stakes domains such as medical diagnostics, autonomous navigation, and assistive technologies, where incorrect yet overconfident predictions can lead to harmful outcomes.

Better-calibrated systems also support more interpretable and human-aligned decision-making, as users can more accurately assess when to rely on model predictions. This contributes to safer human-AI collaboration and accountability, which is increasingly critical as AI systems gain autonomy.





\section*{Theoretical Formulation}

Formally, given a model's softmax outputs $\mathbf{p} = (p_1, \dots, p_K)$ derived from logits $z_i$, the true label $y$, the top predicted confidence $p_{\max} = \max_i p_i$, and the predicted probability of the ground truth class $p_y$, we define the \textbf{soft-calibration loss} as:
\begin{equation}
    \mathcal{L}_{\text{cal}}(p_y, p_{\max}) = p_y(1-p_{\max}) + (1 - p_y)p_{\max}
\end{equation}

The full training objective therefore becomes:
\[
    \mathcal{L}_{\text{tot}} = \mathcal{L}_{\text{FL}} + \lambda \mathcal{L}_{\text{cal}},
\]
where $\lambda$ is a tunable hyperparameter and $\mathcal{L}_{\text{FL}}$ is the Focal Loss~\cite{mukhoti2020calibrating}, which is a strictly proper scoring rule that encourages accuracy and is also found to implicitly enhance calibration. Our calibration term is not an ad-hoc tweak; it arises naturally as a plug-in surrogate to the Upper Bound on Classification Error (UBCE):
\begin{multline*}
    \text{UBCE} = \text{Pr}(t=0)\mathbb{E}\left[p_{\max} \middle| t = 0\right] \\+ \text{Pr}(t=1)\left(1 - \mathbb{E}\left[p_{\max} \middle| t= 1\right]\right),
\end{multline*}
where $t=\mathbb{I}\{\hat{y} = y\}$ is an indicator function for whether the prediction is correct and $\text{Pr}(\cdot)$ defines the probability. Equivalently, algebraic regrouping gives us:
\[
    \text{UBCE} = \mathbb{E}\left[t(1 - p_{\max}) + (1 - t)p_{\max}\right].
\]
This form makes it clear that UBCE is the expected absolute gap between correctness and confidence -- hence a conservative upper bound on calibration metrics. Applying the law of total expectation (tower rule) and conditioning on an input $x$, we have:
\begin{align*}
    \text{UBCE} &= \mathbb{E}_x\left[\mathbb{E}\left[t (1 - p_{\max}) + (1 - t)p_{\max} \middle| x  \right]\right] \\
    &= \mathbb{E}_x\left[\mathbb{E}\left[t \middle| x\right] (1 - p_{\max}) + \mathbb{E}\left[1 - t \middle| x \right]p_{\max} \right].
\end{align*}
Here the outer expectation is over inputs $x$ drawn from the data distribution, and the inner expectation is over the randomness in the correctness indicator $t$ given $x$. Since $p_{\max}$ is deterministic conditioned on $x$, it can be pulled outside the inner expectation.

\begin{table*}[ht!]
  \centering
  \small
  \setlength{\tabcolsep}{1mm}
    \begin{tabular}{l|
                    cccc ccc|
                    cccc ccc}
      \toprule
      {\textbf{Architecture}} 
        & \multicolumn{7}{c|}{\textbf{ScienceQA}} 
        & \multicolumn{7}{c}{\textbf{VQARad}} \\
      \cmidrule(lr){2-8} \cmidrule(l){9-15}
        & $\uparrow$\textbf{Acc.} 
        & $\uparrow$\textbf{F1} 
        & $\uparrow$\textbf{Prec.} 
        & $\uparrow$\textbf{Rec.} 
        & $\downarrow$\textbf{ACE} 
        & $\downarrow$\textbf{ECE} 
        & $\downarrow$\textbf{MCE} 
        & $\uparrow$\textbf{Acc.} 
        & $\uparrow$\textbf{F1} 
        & $\uparrow$\textbf{Prec.} 
        & $\uparrow$\textbf{Rec.} 
        & $\downarrow$\textbf{ACE} 
        & $\downarrow$\textbf{ECE} 
        & $\downarrow$\textbf{MCE} \\
      \midrule
      Granite-vision-3.3-2b
        & 65\%  & 0.467 & 0.501 & 0.442 & 0.439 & 0.343 & 0.75 
        & 59.00\%  & 0.542 & 0.611 & 0.573 & 0.310 & 0.309 & 0.498 \\
      SmolVLM-Instruct  
        & 49\%  & 0.325 & 0.401 & 0.281 & 0.296 & 0.295 & 0.477 & 62.4\%  & 0.62 & 0.623 & 0.62 & \textbf{0.138} & \textbf{0.123} & \textbf{0.224}\\
      InternVL-4B
        & \textbf{84\%}     & 0.569 & 0.585 & 0.557 & \textbf{0.014} & \underline{0.021} & \textbf{0.109}
        & \underline{68.00\%} & \textbf{0.674} & 0.684 & \underline{0.675} & 0.682 & 0.681 & 0.682 \\
      Ristretto-3B
        & 81\%  & \underline{0.652} & \underline{0.664} & \underline{0.641} & \underline{0.017} & \textbf{0.016} & \underline{0.124} & 64.50\%  & 0.636 & 0.648 & 0.638 & 0.647 & 0.645  & 0.651\\
      Ovis2-4B
        & 65\%  & 0.443 & 0.441 & 0.427 & 0.351 & 0.351 & 0.739 & \textbf{69.00\%}  & \underline{0.672} & \textbf{0.730} & \textbf{0.682} & 0.432 & 0.456  & 0.69\\
      H20vl-mississippi-2b
        & 72\%  & 0.470 & 0.474 & 0.47 & 0.172 & 1.722 & 0.6712 & 65.3\%  & 0.644 & \underline{0.692} & 0.665 & 0.653 & 0.654  & 0.656\\
    DeepSeekVL2-Tiny
        & \underline{82\%}  & \textbf{0.696} & \textbf{0.706} & \textbf{0.699} & 0.438 & 0.028 & 0.1741 & 59\%  & 0.49 & 0.602 & 0.548 & \underline{0.28} & \underline{0.128}  & \underline{0.28}\\
      \bottomrule
    \end{tabular}%
    \caption{Comparison of state‐of‐the‐art model performance on the ScienceQA and VQARad datasets. Bold values indicate best performance for each metric within each dataset, underlined values indicate second-best performance.}
    \label{tab:sota_combined_vlm}
\end{table*}

Let $q(x) := \mathbb{E}\left[t \middle| x\right]$ be the true conditional probability of correctness given an input $x$. Then,
\begin{equation*}
    \text{UBCE} = \mathbb{E}_x \left[q(x) (1 - p_{\max}) + (1 - q(x)) p_{\max} \right].
\end{equation*}

In practice, $q(x)$ is unknown because it depends on the non-differentiable indicator function $\mathbb{I}\{\hat{y} = y\}$. To obtain a differentiable surrogate, we use the model's own soft belief about correctness. Instead of taking $\hat{y} = \arg\max_i p_i$, imagine sampling a label $\hat{y} \sim p(\cdot | x)$ from the model's softmax output. Then the expectation of the indicator function $\mathbb{E}\left[t \middle| x\right] = p_y$; that is, the probability that a randomly drawn label equals the true label is exactly $p_y$. This makes $p_y$ a natural, smoothed estimator for $q(x)$. This replacement is an instance of the classical plug-in principle~\cite{plugin_estimators}. Accordingly, we define our soft surrogate loss per example as:
\begin{equation}
    \mathcal{L}_{\text{cal}} (p_y, p_{\max}) = p_y (1 - p_{\max}) + (1 - p_y) p_{\max} \label{eq:l_cal_}
\end{equation}
so that,
\begin{align*}
    \mathbb{E}\left[\mathcal{L}_{\text{cal}}\right] &= \mathbb{E}_x \left[p_y(1-p_{\max}) +(1-p_y)p_{\max}\right] \\
    &\approx \text{UBCE},
\end{align*}
with equality in expectation under the assumption that $p_y \approx \mathbb{E}\left[t \middle| x\right]$. This construction gives a smooth, differentiable surrogate to UBCE that admits gradient-based optimization while preserving the structure of the original bound.

From a probabilistic calibration perspective, if the model is well-calibrated in the sense that its softmax outputs reflect the true posterior---i.e., $p_y \approx \text{Pr}(y|x)$---then $p_y$ also approximates $\text{Pr}(\hat{y} = y | x)$, making the plug-in substitution for $q(x)$ increasingly accurate. There is a self-correcting feedback: early in training $p_y$ might poorly estimate $q(x)$, but the calibration loss penalizes discrepancies between $p_{\max}$ and this current belief. Improving those discrepancies tends to make $p_y$ a more honest estimate of correctness, which in turn makes the surrogate tighter.

\section*{Ablation studies}
 \textbf{Number of Agents} From Table~\ref{tab:agent_calibration} Dropping weaker agents (4→3), improves calibration beyond any individual stage-1 agent, highlighting the debate process’s strength. Among individual agents, the \emph{Knowledge Agent} is least calibrated , while the \emph{CoT Agent} is the strongest single model. 
\begin{table}[ht]
\centering
 \small
\caption{Calibration Metrics by Agent Types}
\label{tab:agent_calibration}
\begin{tabular}{lc|c|c}
\toprule
\textbf{Agent} & \textbf{ACE} & \textbf{ECE} & \textbf{MCE} \\
\midrule
COT Agent                              & 0.1003 & 0.1002 & 0.3164 \\
Knowledge Agent                        & 0.1502 & 0.1616 & 0.6138 \\
Self Ask Agent (SA)                         & 0.1598 & 0.1436 & 0.2529 \\
Search Augmented Agent                 & 0.1140 & 0.1050 & 0.3380 \\
3 Agents (Dropped Knowledge)           & \textbf{0.0919} & 0.0964 & \textbf{0.1604} \\
2 Agents (Dropped Knowledge, SA) & 0.1010 & \textbf{0.0930} & 0.3050 \\
\bottomrule
\end{tabular}
\end{table}
\begin{table}[ht]
\centering
\small
\caption{Calibration Metrics by $\lambda$ (\cal loss)}
\label{tab:lambda_calibration}
\begin{tabular}{lc|c|c}
\toprule
\textbf{$\lambda$} & \textbf{ACE} & \textbf{ECE} & \textbf{MCE} \\
\midrule
1 & 0.1394 & 0.1404 & 0.4147 \\
2 & \textbf{0.1150} & \textbf{0.1370} & 0.4250 \\
3 & 0.1778 & 0.1878 & \textbf{0.3875} \\
4 & 0.1932 & 0.1971 & 0.3971 \\
\bottomrule
\end{tabular}
\end{table}
\\
\\
\noindent\textbf{Ablation of \cal loss (Table \ref{tab:lambda_calibration}), $\lambda$:}  $\lambda=2$ gives the best  calibration 
\\
\\
\noindent \textbf{Debate rounds.} Increasing the number of rounds hurts calibration (ECE $0.146\!\to\!0.1881$, ACE $0.133\!\to\!0.1826$), so we adopt a single round for efficiency; this pattern is consistent with recent findings on test-time scaling \cite{ghosal2025does}
\\
\\
 \textbf{Inference times}: Per question: ~40s (VQA-RAD y/n), 120s (ScienceQA MCQ). Two sequential stages. VQA-RAD: 6.35s + 33.65s; ScienceQA: 9.33s + 110.67s (4 agents). Agents ran sequentially due to GPU limits; parallel agents \& stage pipelining can boost throughput.
\section*{Prompt Template used in Agentic Debate Framework}

\begin{enumerate}

\item \textbf{Stance generation}\\
State your answer (as short as possible, in one or a few words), then rate:
the level of ambiguity in the input query (a float from 0 to 1);
the level of complexity of the input query (a float from 0 to 1);
and your level of ability for solving the input query (a float from 0 to 1).
Note that your uncertainty about correctness is affected by input ambiguity, task complexity, and your own knowledge and abilities.
Based on this, give a float (between 0 and 1) indicating your overall confidence that your answer is correct.
Format: \texttt{Answer: <your short answer>\quad Confidence: <float in [0,1]>}

\item \textbf{Argument generation}\\
You are participating in a debate on the question: “\{QUERY\}”.
Your assigned stance on the question is: “\{STANCE\}”.
Generate arguments or evidence (no more than three sentences) explaining why your assigned stance is correct.
If the question is ambiguous, explicitly state the assumptions or interpretations you adopt.
Be concise and exclude anything irrelevant or unhelpful to supporting the stance.

\item \textbf{Final confidence elicitation}\\
Given the question: “\{QUERY\}”, your original answer is “\{STANCE\}” with a confidence score of \{ORIGINAL-CONFIDENCE\}.
An argument from the opposing side is “\{ARGUMENT\-AGAINST\}”, which received the following rating and feedback from other deliberators: “\{FEEDBACK\-AGAINST\}”. Note that \{NUMBER\_AGAINST\} people disagreed with you.\\
An argument supporting your original answer is “\{ARGUMENT\_FOR\}”, which received the following rating and feedback from other deliberators: “\{FEEDBACK\-SUPPORTING\}”. Note that \{NUMBER\-SUPPORTING\} people agreed with you.\\[4pt]
Provide your final answer to the question (as short as possible). Considering your original belief, group consensus, and these new observations—and weighing arguments from multiple sides (including your own)—give a brief rationale for whether you would adjust your original confidence score.\\
Recall your original confidence is \{ORIGINAL\-CONFIDENCE\}. Given your rationale “\{CONFIDENCE\-RATIONALE\}”, provide your final confidence score (a float in [0,1]) in the exact format: \texttt{Confidence: <float in [0,1]>}.

\end{enumerate}

\end{document}